\documentclass[runningheads]{llncs}
\usepackage[T1]{fontenc}

\usepackage{multirow}
\usepackage{colortbl}
\usepackage{version}
\usepackage{amsmath}
\usepackage{amssymb}
\usepackage{array}
\usepackage{float}
\usepackage{xcolor}
\usepackage{graphicx}
\usepackage{subcaption}
\usepackage{cite}

\graphicspath{{./imgs/}}
\newcolumntype{M}[1]{>{\centering\arraybackslash}m{#1}}
\begin{document}

\title{H-SPAM: Hierarchical Superpixel \\ Anything Model}
\author{Julien Walther$^{\;1, \;2}$ \and Rémi Giraud$^{\;1}$ \and Michaël Clément$^{\;2}$ }
\authorrunning{J. Walther et al.}
\institute{Univ. Bordeaux, CNRS, Bordeaux INP, IMS, UMR 5218, France  \and
Univ. Bordeaux, CNRS, Bordeaux INP, LaBRI, UMR 5800, France
}
\maketitle              

\begin{abstract}
Superpixels offer a compact image representation by grouping pixels into coherent regions.
Recent methods have reached a plateau in terms of segmentation accuracy by generating noisy superpixel shapes.
Moreover, most existing approaches produce a single fixed-scale partition that limits their use in vision pipelines that would benefit multi-scale representations.
In this work, we introduce H-SPAM (Hierarchical Superpixel Anything Model), a unified framework for generating accurate, regular, and perfectly nested hierarchical superpixels.
Starting from a fine partition, guided by deep features and external object priors, H-SPAM constructs the hierarchy through a two-phase region merging process that first preserves object consistency and then allows controlled inter-object grouping. 
The hierarchy can also be modulated using visual attention maps or user input to preserve important regions longer in the hierarchy. 
Experiments on standard benchmarks show that H-SPAM strongly outperforms existing hierarchical methods in both accuracy and regularity, while performing on par with most recent state-of-the-art non-hierarchical methods. 
Code and pretrained models are available: \texttt{\url{https://github.com/waldo-j/hspam}}.
\end{abstract}

\section{Introduction}
\label{sec:intro}
\subsection{Context}

Superpixel decomposition is a standard tool in computer vision. It aims to locally cluster pixels into nearly uniform regions,
leading to a more compact image representation. 
This representation can simplify manual annotation and 
is used in many downstream tasks, including 
semantic segmentation~\cite{mei2025spformer},
transformer-based representation learning~\cite{ke2024learning}, 
implicit neural representations~\cite{li2024superpixel}, 
and 
neural radiance fields~\cite{chen2023structnerf}.

Classical superpixel methods rely on low-level color features.
They are simple to apply to any image, but are mostly driven by local contrasts and struggle to accurately follow semantic object boundaries. 
More recent deep learning-based methods use high-level features to achieve better alignment with image objects. 
Nevertheless, their focus on segmentation accuracy has led them to 
relax the regularity constraint and
produce superpixels with very irregular shapes, 
that are not easily interpretable \cite{giraud2024_tip}.
In both cases, superpixels are typically computed at a single resolution, \emph{i.e.}, for a required number of superpixels, which limits their use for tasks that would benefit from a full hierarchical segmentation, such as interactive annotation 
\cite{berg2019ilastik}
or multi-scale reasoning~\cite{2025HieraASGSeg}.
Hierarchical variants exist for both
traditional, \emph{e.g.} \cite{chang2024hierarchical}
and deep approaches \emph{e.g.} \cite{peng2021superpixelsdeepaffinitylearning}. 
However, similarly to non-hierarchical methods, their segmentation accuracy is generally improved by generating noisy superpixel shapes, to try to capture object boundaries. 

Notably, recent backbone-level grouping methods keep revisiting how to build semantic segmentation hierarchies, from superpixel-based progressive graph pooling in~\cite{ke2024learning} to content-aware token grouping that yields native multi-scale masks in~\cite{braso2025native}, showing that hierarchical construction remains an active research direction.
These developments underline the growing interest in multi-scale, hierarchical segmentation in both supervised and unsupervised frameworks.
At the same time, segmentation foundation models such as SAM~\cite{kirillov23sam} provide a 
deep understanding of object boundaries, regardless of their semantic. SPAM~\cite{Walther2025spam} recently showed the benefit of using such object priors to achieve higher superpixel segmentation performance, yet it is not a hierarchical model.
Therefore, we introduce in this paper a unified hierarchical framework that produces accurate and regular superpixels, 
that are perfectly nested across all scales.

\subsection{Related Works}

\noindent\textbf{Traditional methods.}  
Early superpixel segmentation approaches
rely on handcrafted low-level features.
Superpixel decomposition were made popular with the introduction of SLIC~\cite{achanta2012} 
that formulates the problem as a $k$-means clustering in the joint color–spatial domain, offering a practical balance between superpixel compactness and accuracy.
Later works improved accuracy through
block refinement \cite{vandenbergh2012},
extended low-level space \cite{chen2017},
or
non-iterative clustering \cite{achanta2017superpixels}.
Many approaches were proposed over the years, yet they all remain limited in terms of segmentation accuracy by their similarity criteria based on low-level features. 
We refer the reader to~\cite{stutz2018superpixels} for an in depth review of traditional superpixel methods.

\noindent\textbf{Deep learning-based methods.}  
The rise of deep learning brought a major shift in superpixel segmentation.
Instead of relying solely on color and spatial proximity, neural approaches learn high-level representations that better align with 
object boundaries.
For example,~\cite{jampani2018superpixel} introduced a differentiable $k$-means clustering clustering scheme, allowing networks to be trained end-to-end for superpixel generation.
Subsequent works proposed variants incorporating contour losses~\cite{tu2018learning}, encoder–decoder networks~\cite{yang2020superpixel}, or neighborhood embeddings~\cite{wang2021ainet}, leading to better alignment with object boundaries.
However, these methods often relax compactness and regularity constraints to maximize accuracy, producing irregular and even sometimes fragmented superpixels.
As highlighted in several evaluations~\cite{stutz2018superpixels,giraud2024_tip}, regularity remains an essential criterion for interpretability and for downstream tasks such as graph-based modeling~\cite{gould2008}, video tracking~\cite{chang2013video}, or interactive segmentation~\cite{zhou2023vine}.

\begin{figure}[t]
\centering
{\scriptsize
\begin{tabular}{ccc}
\includegraphics[width=0.32\linewidth]{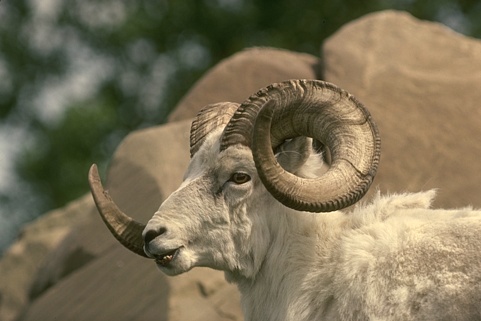} &
\includegraphics[width=0.32\linewidth]{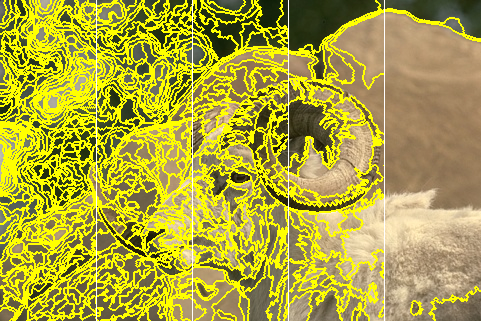} &
\includegraphics[width=0.32\linewidth]{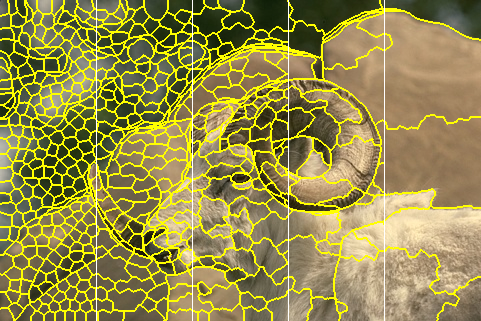}\\
(a) Image &(b) HHTS \cite{chang2024hierarchical} (CVPR'24)&
(c) H-SPAM (ours) \\[-1ex]
\end{tabular}}
\caption{\textbf{Hierarchical superpixel segmentation example.} The proposed H-SPAM method generates very regular and easily interpretable superpixels that are aligned with the image objects through the hierarchy, compared to HHTS \cite{chang2024hierarchical}.}\vspace{-0.25cm}
\label{fig:teaser}
\end{figure}

\noindent\textbf{Hierarchical superpixels.}  
While progress has been made in accuracy and regularity, most existing methods operate at a single scale.
Hierarchical decompositions solve this by enforcing nestedness: boundaries of a finer scale are fully contained within those of the coarser one. This is useful for progressive refinement, efficient annotation, and multi-level reasoning.
Some works build this type of hierarchy with region merging on graph, for example with minimum spanning tree strategy~\cite{wei2018} or iterative graph merging~\cite{galvao2020image}. More recent algorithms rely on information-based grouping~\cite{xie2025hierarchical}, and some use a neural network to learn affinities before constructing the hierarchy~\cite{peng2021superpixelsdeepaffinitylearning}, or with a focus on efficiency~\cite{yan2022hierarchical}.
However, ensuring nestedness across scales necessarily reduces segmentation accuracy and contributes to making superpixels less regular as can be seen in Figure~\ref{fig:teaser}(b). This highlights the need of a hierarchical method that is both accurate and regular.

\noindent\textbf{Object-based superpixels.}  
Recently, the Superpixel Anything Model framework (SPAM)~\cite{Walther2025spam} showed that guiding superpixels with high-level object masks from large pretrained segmentation models such as SAM~\cite{kirillov23sam} leads to very accurate and regular decompositions. 
Using such priors enables to 
overcome the performance plateau on natural images of 
state-of-the-art superpixel methods. 
High-level object masks and low-level superpixels complement each other: object priors provide semantic structure and help clarify ambiguous regions, while superpixels offer a finer and regular partition that can capture object subparts and supports downstream vision tasks.
However, this approach only provides a single level of over-segmentation.

\subsection{Contributions}

In this work, we introduce a hierarchical method called H-SPAM (Hierarchical Superpixel Anything Model), 
which unifies the accuracy of object-based superpixels with the benefits of multi-scale segmentation.
By using prior object masks, our method enforces 
boundary alignment while maintaining perfect hierarchical consistency across scales (see Figure \ref{fig:teaser}(c)). Our main contributions are:
\begin{itemize}
    \item \textbf{Object-based hierarchical segmentation:} The first object-based deep learning framework for hierarchical superpixels integrating prior object masks. H-SPAM establishes a new benchmark for hierarchical superpixel decomposition by combining multi-scale consistency with high segmentation accuracy and regularity.
    This is achieved through a two-phase process that first performs strict intra-object merging guided by object priors, and then allows inter-object merges to build a complete hierarchy across scales.
    
    \item \textbf{Adaptive modes:} We propose two 
    segmentation modes using 
    visual attention. A visual attention map or user inputs (clicks) can be used to preserve more superpixels in objects of interest through the hierarchy. 
    These modes offer complementary alternatives to the multi-scale superpixel segmentation.
    \item \textbf{Extensive evaluation:} We evaluate our method on standard datasets and relevant metrics, comparing it to the state-of-the-art methods. 
    H-SPAM offers a perfectly hierarchical decomposition, and largely outperforms other hierarchical methods in terms of both accuracy and regularity, while maintaining the performance of single scale object-based approaches~\cite{Walther2025spam}.
\end{itemize}

\begin{figure}[t]
	\centering
	\includegraphics[width=0.99\linewidth]{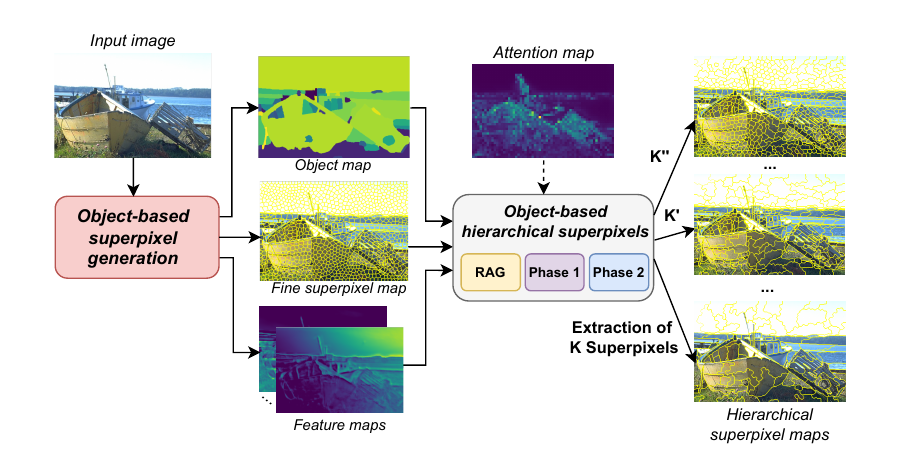} \\[-2.5ex]
	\caption{ \textbf{Global framework of the H-SPAM method.} An object-based superpixel segmentation provides a fine superpixel map along with a high-level object map and features that are used by to create a hierarchy that respects the image objects through the merging process.
    Visual attention can also be used to concentrate or delay the merge on specific areas.}
	\label{fig:framework}
\end{figure}

\section{Hierarchical Superpixel Anything Model}

In this Section, we present a fast and simple hierarchical region merging method, illustrated in Figure~\ref{fig:framework}, which builds a segmentation hierarchy from i) an initial fine superpixel partition, ii) deep features, iii) a prior object map and iv) an optional visual attention signals to guide the construction of the hierarchy.
The algorithm follows a two-phase strategy: it first merges regions inside objects of the prior map only, then progressively merges objects. The resulting tree represents the scene at multiple levels of granularity while offering explicit control over the number of generated regions and their regularity.

\subsection{Input Features and Prior Superpixel Partition}

The first stage of our framework is designed to extract deep feature representations, generate an initial fine-scale superpixel partition, and produce a corresponding object map.
These are generated using \cite{Walther2025spam}, which operates as a deep superpixel segmentation framework built upon a convolutional neural network (CNN) and a differentiable clustering module.
Given an input image $I$ and a prior object segmentation computed by Segment Anything Model (SAM)~\cite{kirillov23sam}, the model jointly produces a dense, fine-grained decomposition of the image into very accurate superpixels.

The model used to get the fine superpixel map generates $d=20$ features.
Among these features, there are 3 channels representing the input LAB image ($F_C$), 2 channels representing the 2D pixel positions ($F_S$), 
then the remaining channels ($d-5$) are deep features ($F_D$) obtained with the convolutional model of \cite{Walther2025spam}.
The combined feature vector is denoted as: 
$\mathbf{F} = 
[
  F_C\in \mathbb{R}^{3},
  F_S\in \mathbb{R}^{2},
  F_D\in \mathbb{R}^{d-5}
] \in \mathbb{R}^{d} .$
The associated object map is made of $M$ different objects denoted $O = \{O_1, \ldots, O_{M}\}$.

In this work, we leverage the same features $\mathbf{F}$ learned for the clustering process to guide the construction of the superpixel hierarchy.
These features capture both semantic and spatial components, allowing the hierarchical representation to remain consistent across scales while preserving object boundaries and fine structural details.

\subsection{Object-based Hierarchical Superpixels}

Our objective is to build a full segmentation hierarchy by progressively merging adjacent regions starting from the initial fine superpixel map, \emph{e.g.} as in \cite{galvao2020image}, while keeping the prior objects as long as possible in the hierarchy.
Following the intuition behind 
Agglomerative Hierarchical Clustering (AHC),
we formulate our hierarchical algorithm with a Region Adjacency Graph (RAG) and iteratively select the pair of neighboring regions with the lowest merging cost.
Thus, each merge defines the next coarser level while preserving a perfectly nested structure across scales.

\begin{figure}[t]
\centering
{\scriptsize
\begin{tabular}{cc}
\includegraphics[width=0.39\linewidth]{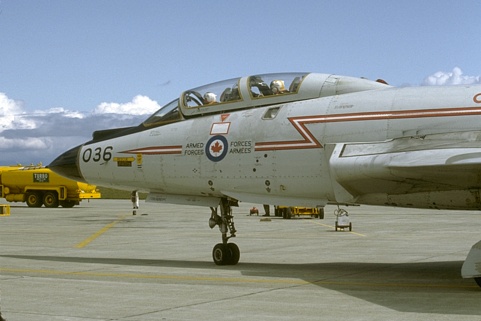} &
\includegraphics[width=0.59\linewidth]{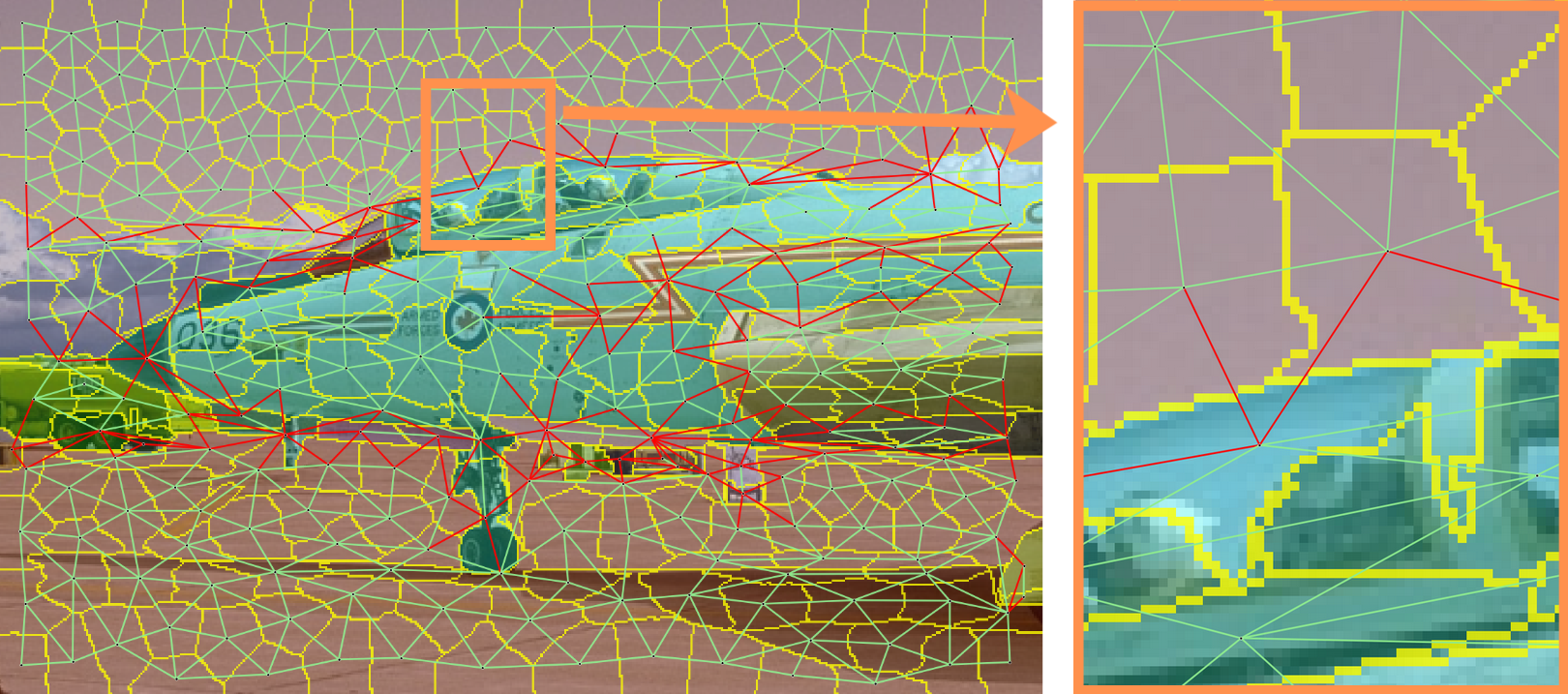} \\

(a) Image &
(b) Object-based Superpixel RAG 
\end{tabular}
}
\caption{\textbf{Object-based superpixel region adjacency graph}. (b) Superposition of the prior object map, the initial superpixels, and the edges between the superpixels. The red ({\color{red}\textbf{--}}) and green  ({\color{green}\textbf{--}}) connections respectively mean that the superpixels do not belong or belong to the same object.}
\label{fig:RAG}
\end{figure}

\subsubsection{Object-based region adjacency graph.}
We first construct a fine superpixel segmentation from the original image $I$. We denote this set as $S = \{S_1, \ldots, S_{N_{f}}\}$, where $N_{f}$ is the number of superpixels in the starting fine segmentation.
We then construct the initial RAG from these superpixels, with V representing the nodes and E representing the edges, defined as:
\begin{equation}
    G^{N_{f}} = (V^{N_{f}},E^{N_{f}}), \quad V^{N_{f}} = \{1, \ldots, N_{f}\}, \quad 
E^{N_{f}} = \{(u,v) \mid S_u \sim S_v \},
\label{eq:graph_def}
\end{equation}
\noindent where $S_u \sim S_v$ indicates that superpixels $S_u$ and $S_v$ share a boundary in the image. So $E$ is the set of all pairs of superpixels that share a boundary.

Let $G^{N_f}$ denote the finest-level graph. Our goal is to construct the full sequence of graphs $\{G^s\}_{1 \le s \le N_f}$ that defines the hierarchy, where each level $s$ corresponds to a graph $G^{s}$. The construction forms a true hierarchy if, for any two levels $i<j$, the boundaries represented at level $i$ are entirely contained within those at level $j$.

Each region $S_u$ at each step is represented by a feature vector $\boldsymbol{\mu}_u \in \mathbb{R}^d$, obtained by averaging over all the pixel features $\mathbf{F}(S_u)$ of the superpixel. The size, \textit{i.e.} the number of pixels in $S_u$  is denoted $s_u = |S_u|$.

Thus, we initialize the graph to distinguish edges between adjacent regions that share the same object prior from those that do not, as shown in Figure~\ref{fig:RAG}(b) with green (red) for same (different) objects.

\subsubsection{Hierarchical superpixel merging.}
The hierarchical merging process is driven by a strategy based on priority queues. 
Specifically, at scale $s$, we select the edge $(u,v) \in E^{(s)}$ that has the lower cost and merge the corresponding adjacent regions into a new region $w$: $S_{w} = S_u \cup S_v$ | $w \notin V^{(s)}$. The node and edge sets are then updated as:
\begin{equation}
V^{(s-1)} = \big( V^{(s)} \setminus \{u,v\} \big) \cup \{w\}, \qquad
E^{(s-1)} = \big( E^{(s)} \setminus (\mathcal{N}_u \cup \mathcal{N}_v) \big) \cup \mathcal{N}_w ,
\end{equation}

\noindent where $\mathcal{N}_u $ and $\mathcal{N}_v$ denote the sets of edges of respectively $u$ and $v$, and $\mathcal{N}_w = \{(w,x)\,|\, x\in V^{(s-1)},\, S_x \sim S_w\}$ collects the edges between the new region and its neighbors.

After each fusion, the features of the newly formed region are updated as a weighted average of the merged regions $\mu_{w} = \frac{s_u \cdot \mu_u + s_v \cdot \mu_v}{s_u + s_v}$, while its size is obtained by summing their respective sizes $s_{w} = s_u + s_v $.

\smallskip

\begin{figure}[t]
	\centering
	\includegraphics[width=\linewidth]{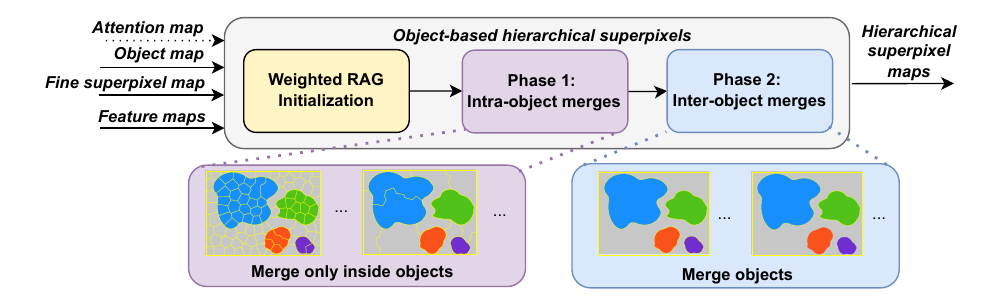}
	\caption{\textbf{Object-based hierarchy creation}. Our merging process combines 2 distinct phases corresponding to the processing of high-level prior objects. The first one only allow merges inside objects while the second one merges objects.}
	\label{fig:hierarchy_creation}
\end{figure}

Our object-based hierarchical clustering process is illustrated in Figure \ref{fig:hierarchy_creation}.
At each phase, the edge of minimum cost is extracted, and the corresponding regions are merged into a new one.
Similarly to~\cite{galvao2020image}, 
our method identifies two regimes and
is consequently designed in two phases, adapted to each one.
The first phase focuses on merging superpixels that share the same object, while the second merges objects together. 
The idea is to preserve the object boundaries as long as possible while
accurately merging sub-objects using the deep features.

\paragraph{Phase 1: Intra-object merges.}
Let $\Theta : V \rightarrow \{1,\dots,M\}$ denote the function assigning each superpixel $u \in V$ to its object. 
During this first phase, only regions that share the same object prior $\Theta(u) = \Theta(v)$ are eligible for merging, which is labelled "Intra-object merges" in Figure~\ref{fig:hierarchy_creation}. The goal is to progressively refine the segmentation inside each object, so that early fusions strictly respect the prior partition. 
To ensure this across all scales of the hierarchy, we propose to set a hard constraint between superpixels that are not included in the same object.
In practice, this is enforced by assigning an infinite cost to all inter-object connections, so no merge can occur between different objects during phase 1. 

\noindent Then $\Theta(u) \neq \Theta(v) \;\Longrightarrow\; \operatorname{cost}(u,v) = +\infty.$

Among all admissible pairs $(u,v)$ with $\Theta(u) = \Theta(v)$, the merging priority is driven by a feature-based cost that combines appearance and spatial terms:
\begin{equation}
\text{$cost_{phase1}$}(u,v,s) = \left\|
\mu_u^{{\{F_C, F_D\}}} - \mu_v^{{\{F_C, F_D\}}} \right\|
_2^2 +  w^{\text{(s)}} \left\| \mu_u^{\{F_S\}} - \mu_v^{\{F_S\}} \right\|_2^2,
\label{eq:cost_step1}
\end{equation}

\noindent where $w^{\text{(s)}}= w_{pos}.\sqrt{\frac{s}{N_{f}}}$. The coefficient $w_{pos}$ controls the importance of the spatial features, \emph{i.e.}, offers direct control over the regularity of the segmentation.
Similarly to 
the SLIC distance \cite{achanta2012}, we automatically adjust the importance of the spatial cost according to the scale, here the hierarchy step.
Indeed, as regions become larger, the spatial distance between regions also increases.

At each merge, the features of the new region are updated as the size-weighted average of the features of the two merged regions. The sequence of merges obtained at this stage defines a finer intra-object hierarchy that remains fully consistent with the object priors.

\paragraph{Phase 2: Inter-object merges.}
Once intra-object merges are completed, each object has been reduced to a single region. The process then switches to the second phase, where regions from different objects are allowed to merge ($\Theta(u) \neq \Theta(v)$). This corresponds to phase 2, where the remaining inter-object (red edges in Figure~\ref{fig:hierarchy_creation}) are progressively merged.

In this phase,
inspired by the Ward's method,  
the cost is modified to explicitly take region sizes into account: \vspace{-0.1cm}
\begin{equation}
\text{$cost_{phase2}$}(u,v) = \frac{s_u.s_v}{s_u + s_v}\Big\lVert \mu_u^{\{F_C,F_D\}} - \mu_v^{\{F_C,F_D\}} \Big\rVert_2^2,
\label{eq:cost_step2}
\end{equation}
so that the priority of inter-object merges depends both on feature similarity and on the sizes $s_u$ and $s_v$ of the regions. 
This factor tends to merge small objects together, then large with small ones, and finally large objects. 
Since only high-level objects remain and we already use a factor considering object sizes, we no longer include the spatial features $F_{S}$ in the cost.
All regions are merged 
until only one connected component remains. 
Thus, this second phase lets the hierarchy evolve beyond the prior object partition and span the full range from the initial superpixels down to a single region.
Overall, our method produces an ordered sequence of merges $\mathcal{M} = \{(u_1,v_1), \ldots, (u_{N_{f}},v_{N_{f}})\},$
defining a hierarchical tree structure over the initial superpixels. With our Python implementation, the hierarchy is obtained in 0.44s for a 481x321 image and for 1250 levels.

\medskip\noindent\textbf{Extraction of intermediate partitions.}
The merge sequence enables the reconstruction of partitions at arbitrary levels of granularity. To obtain exactly $K$ regions, the first $N_{f}-K$ merges of $\mathcal{M}$ are applied. This is done in only $0.07$s on average. Thus, for any $K \in [1,N_{f}]$, the method yields a consistent hierarchical segmentation, preserving the history of merges while enabling extraction of partitions at fixed resolution.

\subsection{Attention-based Modulation}

In H-SPAM, we propose alternative modes using visual attention to adjust the distribution of superpixels in the regions of interest.
Similarly to \cite{Walther2025spam}, the underlying idea is to allocate a higher density of superpixels to areas containing meaningful sub-objects, while using fewer regions in smoother or less relevant parts of the image, such as the background. 
This non-uniform allocation is intended to better support some downstream tasks, for instance annotation, without enforcing a uniform level of detail over the entire image.
We set $a_u$ as the mean pixel-wise attention inside a superpixel $u$.
Without using object priors, this mode only relies on a coarse attention map that can be imprecise ("w/o objects" in Figure~\ref{fig:Va_attention}). 
Instead, we propose to average attention within each high-level object ("w/ objects" in Figure~\ref{fig:Va_attention}), so $a_u$ is the mean attention over $\Theta(u)$, yielding more coherent scores.

When this option is enabled, the fusion cost between two adjacent superpixels $u$ and $v$ is updated as: \vspace{-0.1cm}
\begin{equation}
\text{cost}(u,v) \leftarrow \text{cost}(u,v) + w_{att} \cdot g(a_u, a_v),
\label{eq:attention_cost}
\end{equation}
where $w_{att} > 0$ weights the modulation and $g(a_u,a_v)=\max(a_u,a_v)$ by default.
We use the maximum to reflect the intuition that if an object contains a highly attended region, the object should be preserved as much as possible in the hierarchy.
This makes the hierarchy sensitive to semantic cues: merges of
salient superpixels are delayed 
while less salient regions merge earlier. 

 With standard hierarchical clustering, 
 visually meaningful or small salient regions may be merged too early, particularly when boundaries are weak.
 In our case, the object prior map 
 from~\cite{kirillov23sam} 
 may also fail to capture them reliably.
By incorporating attention, semantically important superpixels are protected from premature merging, remaining isolated until higher hierarchy levels and improving the recovery of fine or thin structures.

\begin{figure}[t]
\newcommand{\hhh}{0.225\textwidth}
\centering
{\scriptsize
\begin{tabular}{cccccc}
&\rotatebox{90}{\textbf{\text{ }w/o objects}}
&\includegraphics[width=\hhh]{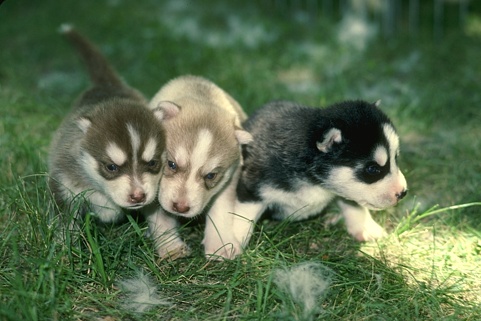} &
\includegraphics[width=\hhh]{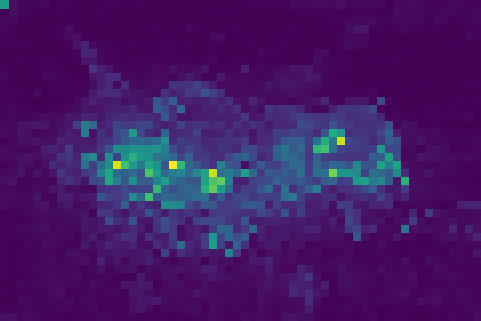} &
\includegraphics[width=\hhh]{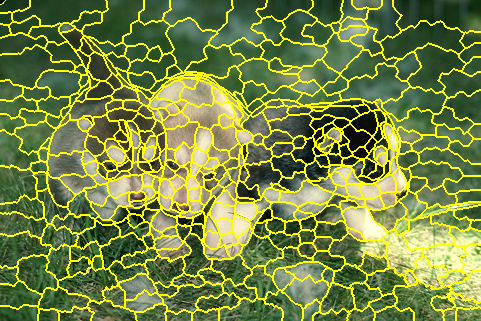} &
\includegraphics[width=\hhh]{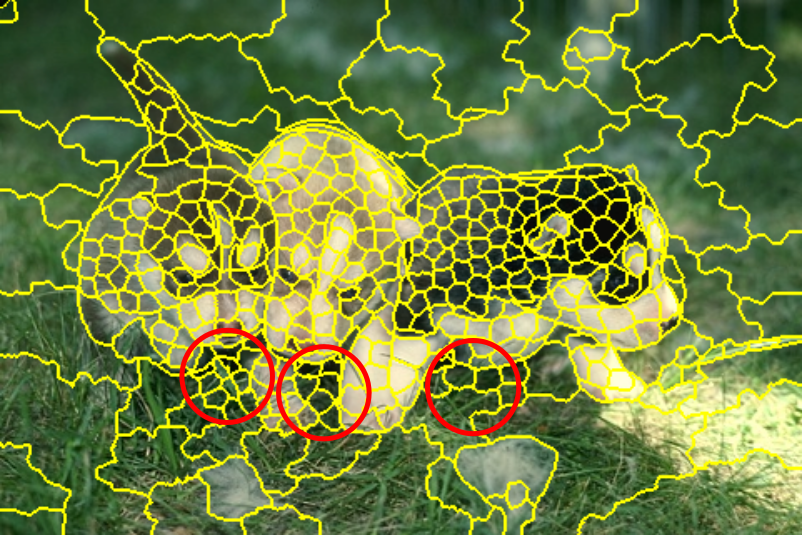} \\
&\rotatebox{90}{\textbf{\text{ }w/ objects}}
&\includegraphics[width=\hhh]{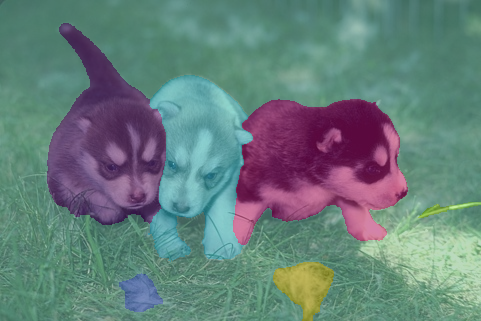} &
\includegraphics[width=\hhh]{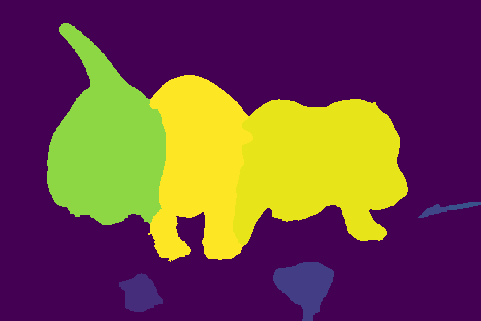} &
\includegraphics[width=\hhh]{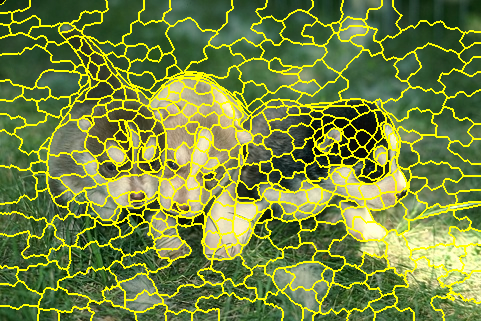} &
\includegraphics[width=\hhh]{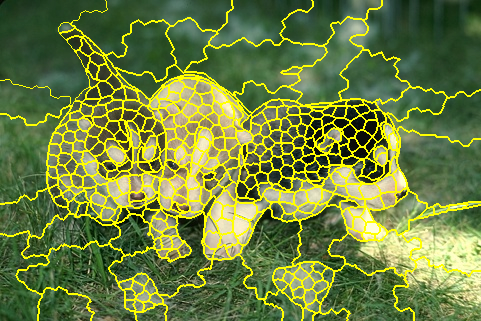} \\
\rotatebox{90}{\textbf{\text{ }w/ objects}}&\rotatebox{90}{\textbf{\text{ }\text{ }\text{ }+clicks}}
&\includegraphics[width=\hhh]{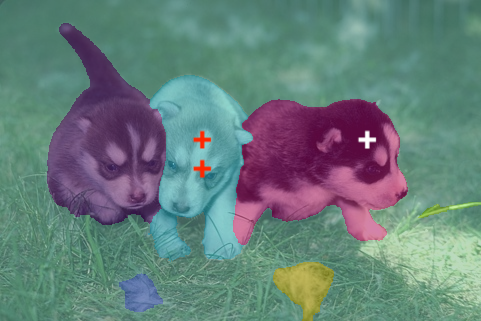} &
\includegraphics[width=\hhh]{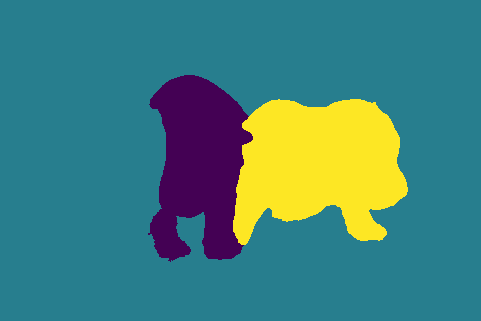} &
\includegraphics[width=\hhh]{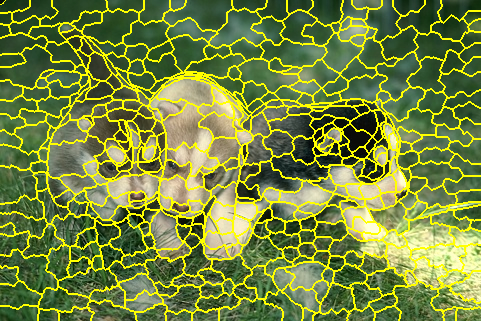} &
\includegraphics[width=\hhh]{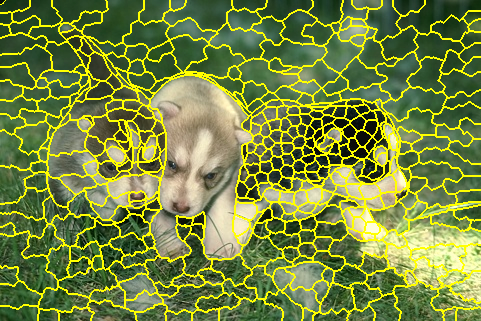} \\
&&Object overlay &
Attention map &
$w_{att}=0.01$ &
$w_{att}=0.5$\\[-1ex]
\end{tabular}}
\caption{\textbf{Illustration of the attention modes}. With our object-based framework (middle row), the attention can be averaged within objects to provide a cleaner guide for the merging process. The bottom row shows the user interactive mode, where red/white crosses lead to fewer/more superpixels in an object, offering a complementary alternative to our multi-scale segmentation method.}
\label{fig:Va_attention}
\end{figure}

The attention maps used in our experiments are obtained from DINO~\cite{caron2021emerging}, but any model or user-provided map can be used. This flexibility enables adjusting the hierarchy to various sources of semantic information, ranging from automated saliency to interactive refinement.
In Figure~\ref{fig:Va_attention},
we compare the different modes using attention, with two different $w_{att}$.
The first row
shows the use of attention without using an object map. 
Attention is averaged over the superpixels,
leading to a focus of resolution over certain regions but with clear leakage artefacts 
around the dog paws (see red circles). 
In the second row, our object-based approach enables to average the attention over the prior objects, preventing such leakage.
 The bottom row shows the interactive mode, where the user has full control to less or more preserve superpixels in objects. 
 
\section{Experimental Results}
\label{sec:results}

In this Section we present qualitative and quantitative evaluation of H-SPAM,
on standard datasets and using recommended metrics \cite{giraud2024_tip}. 
We report the impact of our method parameters and comparison to both hierarchical and non-hierarchical state-of-the-art superpixel methods.

\subsection{Validation Framework}
\label{subsec:validation}

\noindent \textbf{Datasets.}
We evaluate on standard superpixel benchmarks that provide manual semantic-agnostic precise object annotations: BSD~\cite{martin2001}, 
NYUv2~\cite{silberman2012indoor} and SBD~\cite{gould2009decomposing} with respectively 200, 399 and 477 test images. Note that the BSD provides multiple groundtruth segmentations per image and also 200 train and 100 validation images that deep learning-based methods use to train their model.

\smallskip

\noindent\textbf{Evaluation metrics.}
Following~\cite{stutz2018superpixels,giraud2024_tip}, we report the main criteria for superpixel quality: Achievable Segmentation Accuracy (ASA) to measure the adherence to objects and Boundary Recall (BR) for contour alignment.
BR is compared to Contour Density (CD)~\cite{machairas2015}, measuring the number of superpixel boundaries.
We measure the regularity with the Shape Regularity Criteria (SRC)~\cite{giraud2017_jei}, that evaluates the convexity, contour smoothness and 2D balance of each superpixel shape. 
SRC was proposed to address the limitations of circularity, often used to evaluate superpixel regularity \cite{giraud2017_jei}.
The compared 
state-of-the-art superpixel methods that we considered hierarchical 
\cite{wei2018,galvao2020image,yan2022hierarchical,chang2024hierarchical,xie2025hierarchical}, are the ones that obtained a perfect Nestedness score of 1~\cite{belem2024measuring}, \emph{i.e.} 
whose regions for a given scale are perfectly contained into the ones of a higher scale.

\subsection{Ablation Study}
\label{subsec:ablation}
\subsubsection{Influence of object prior.}

\begin{figure}[t]
\centering
{\scriptsize
\begin{tabular}{cccc}
\includegraphics[width=0.24\textwidth,height=0.16\textwidth]{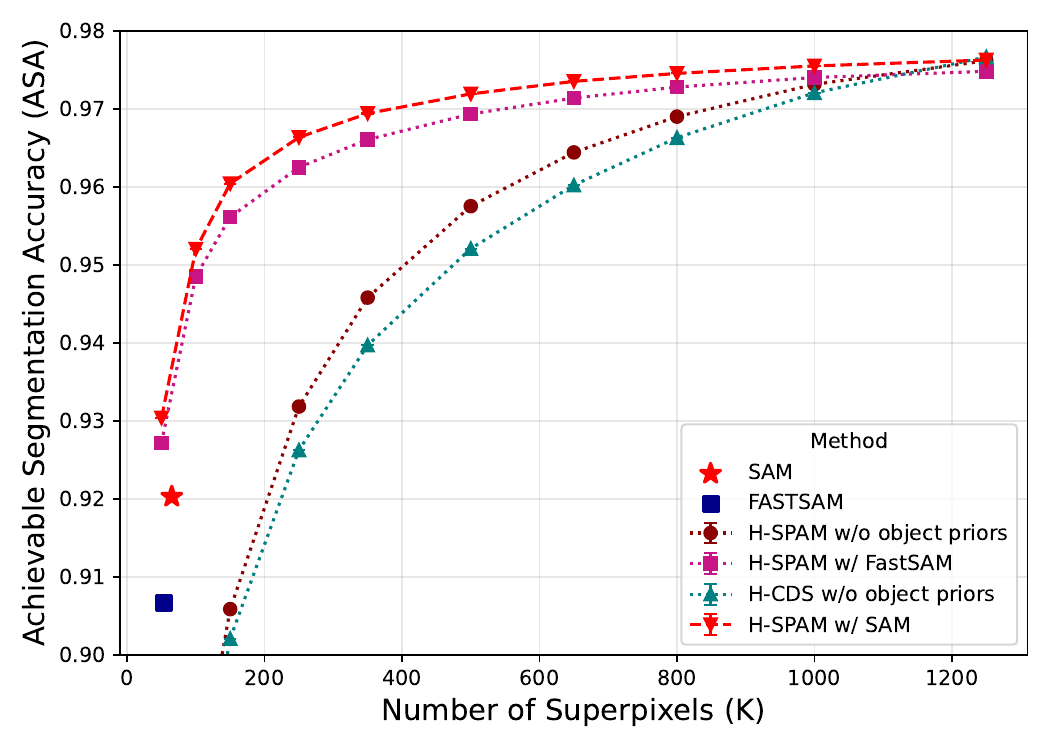} &
\includegraphics[width=0.24\textwidth,height=0.16\textwidth]{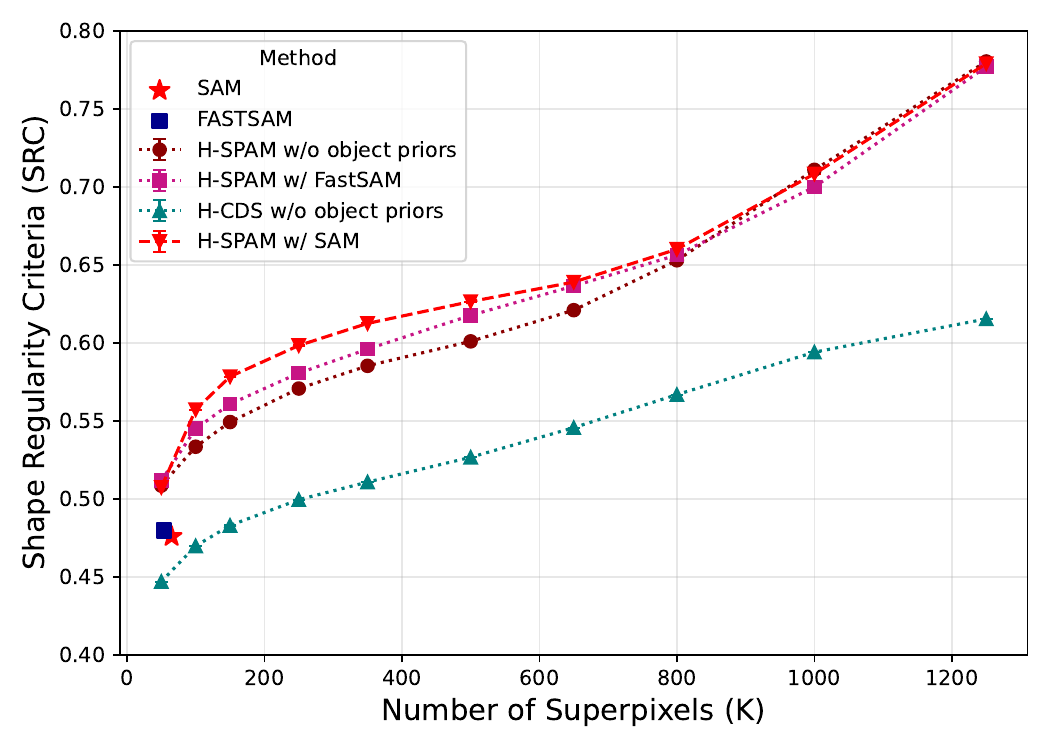} &
\includegraphics[width=0.24\textwidth,height=0.16\textwidth]{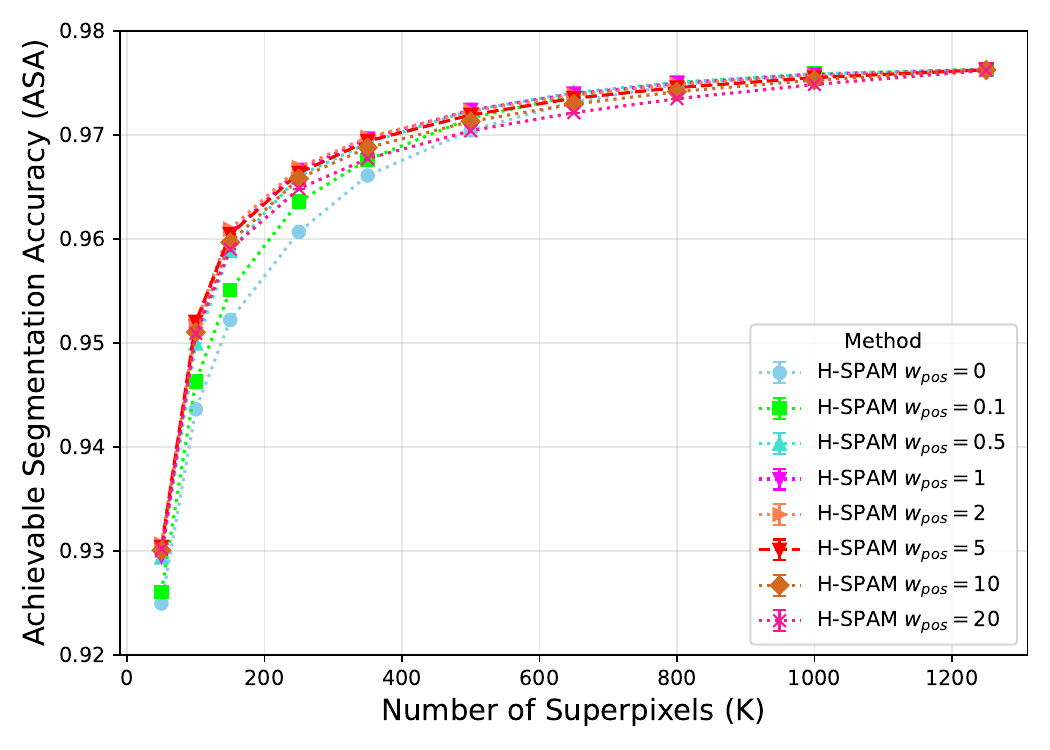} &
\includegraphics[width=0.24\textwidth,height=0.16\textwidth]{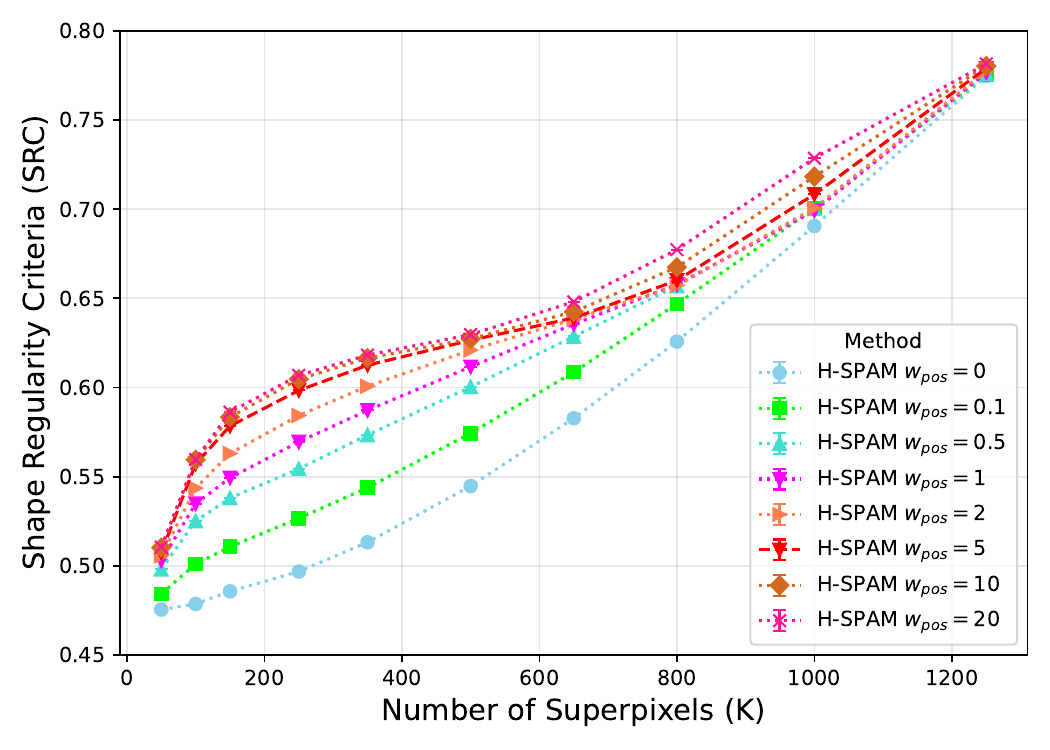} \\
\multicolumn{2}{c}{(a) Impact of object prior}  &
\multicolumn{2}{c}{(b) Impact of $w_{pos}$} \\[-1ex]
\end{tabular}
}
\caption{\textbf{Ablation study - Influence of parameters.} (a) Importance of the prior mask. Using object priors to guide the hierarchy largely improves the segmentation accuracy.
(b) Influence of $w_{pos}$ on the accuracy and regularity.}
\label{fig:Ablation_metrics}
\end{figure}

Figure~\ref{fig:Ablation_metrics}(a) compares H-SPAM with and without high-level object map priors during hierarchy construction. 
When used, the object map is the same that 
constrained the fine superpixel partition.
Including object prior constraints again to build the hierarchy helps to guide the merging within objects and to limit cross-object merging.
The segmentation accuracy (ASA) is largely improved, especially at low scale, \emph{i.e.}, low $K$.
We compared the use of SAM \cite{kirillov23sam} and also FastSAM~\cite{zhao2023fast}
to generate the fine partition and constrain the merging.
FastSAM reduces computation time,
with moderate impact of accuracy and regularity (SRC).

We also reported the mean performance obtained using only high-level objects from SAM and FastSAM. 
The significant gap of performance shows
that the prior object map is very helpful to guide the segmentation and merging but not sufficient to capture all image objects. 
This demonstrates the interest of such object-based approaches, generating finer superpixels within objects.

Finally, 
we also evaluate our hierarchy creation method on fine superpixel maps from CDS \cite{xu2024learning}.
These maps obtain similar accuracy to the ones of SPAM \cite{Walther2025spam} for high $K$,
but are not constrained by object priors,
and are much less regular.
This variant (H-CDS) obtains inferior segmentation accuracy than H-SPAM w/o object priors.
This demonstrates 
the usefulness of our approach using an object-based fine superpixel partition
and that regularity may play a key role in the hierarchy creation to preserve object boundaries.

\begin{figure}[t]
    \centering
    {\footnotesize
    \begin{tabular}{cccc}
    \includegraphics[width=0.24\textwidth]{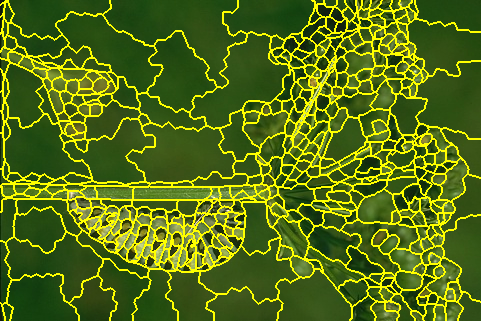} &
    \includegraphics[width=0.24\textwidth]{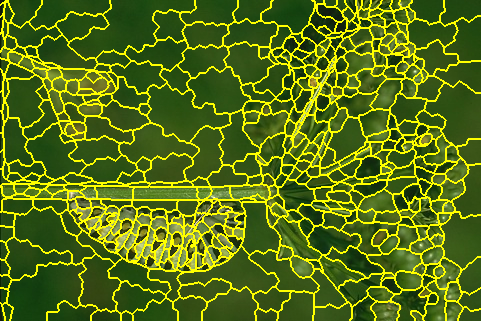} &
    \includegraphics[width=0.24\textwidth]{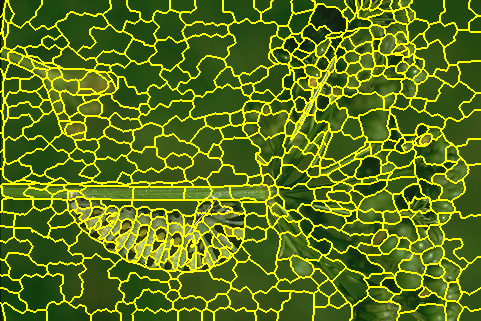} &
    \includegraphics[width=0.24\textwidth]{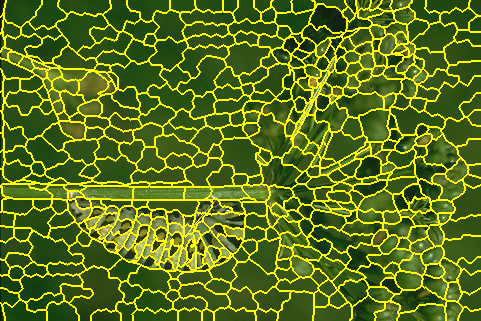} \\
    $w_{pos}=0.1$ &
    $w_{pos}=1$ &
    $w_{pos}=5$ &
    $w_{pos}=10$ \\[-1.5ex]
    \end{tabular}
    }
	\caption{\textbf{Influence of the spatial parameter $w_{pos}$} for K=500 superpixels. 
    This parameter controls the shape consistency of the decomposition. 
    A high $w_{pos}$ enforces smallest shapes to merge.
    By default, we use $w_{pos}=5$.    }
	\label{fig:images_wpos}
\end{figure}

\noindent\textbf{Influence of the parameter $w_{pos}$.}
The parameter $w_{pos}$ controls the importance of spatial features, thus 
adjusts the regularity of 
superpixels during the construction of the hierarchy (see Figure \ref{fig:images_wpos}).
Figure \ref{fig:Ablation_metrics}(b) shows the influence of this parameter on ASA and 
SRC. 
As expected, high $w_{pos}$ lead to more regular superpixels, but low $w_{pos}$ also lead to reduced accuracy.
While reducing the regularity generally helps superpixels to capture image objects, 
region consistency seems to help the hierarchy to preserve object boundaries.
Therefore, we chose an intermediate $w_{pos}=5$ as default value in our experiments.

\subsection{Quantitative evaluation}

\begin{figure*}[t!]
\centering
\newcommand{\heee}{0.2075\textwidth}
{\scriptsize
 \begin{tabular}{@{\hspace{0mm}}c@{\hspace{2mm}}c@{\hspace{2mm}}c@{\hspace{2mm}}c@{\hspace{2mm}}c@{\hspace{0mm}}}

 & \textbf{ASA} & \textbf{SRC} & \textbf{BR}  \\
 \rotatebox{90}{\hspace{1cm}{\textbf{BSD}}}&
\includegraphics[width=0.31\textwidth, height=\heee]{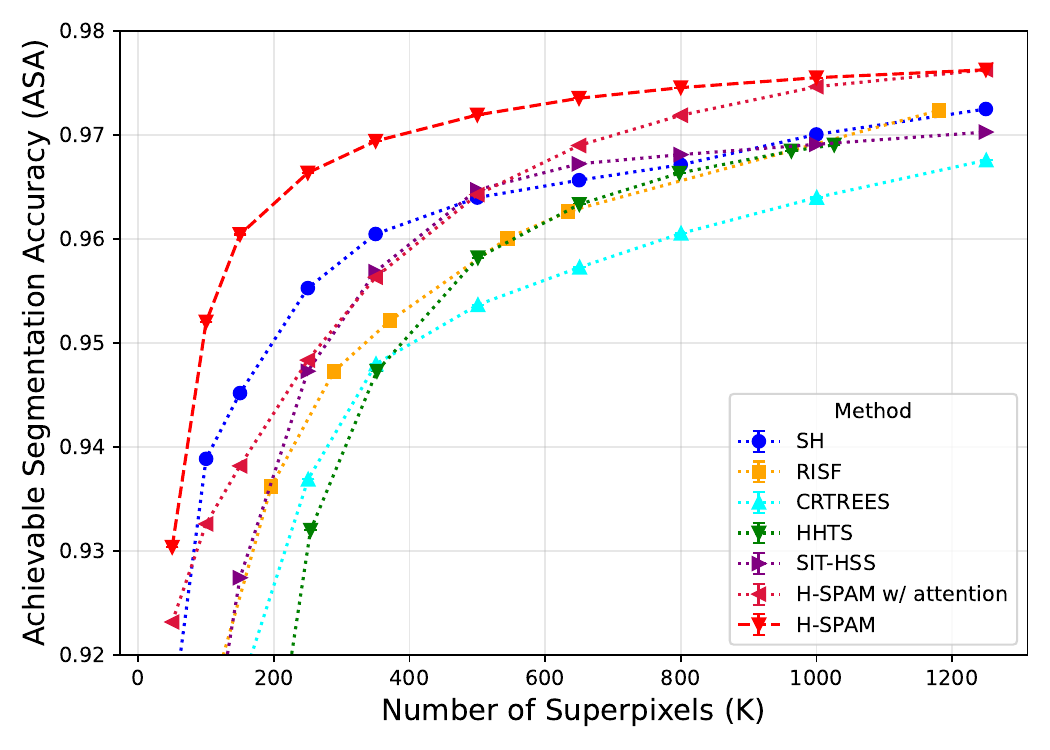}&
\includegraphics[width=0.31\textwidth, height=\heee]{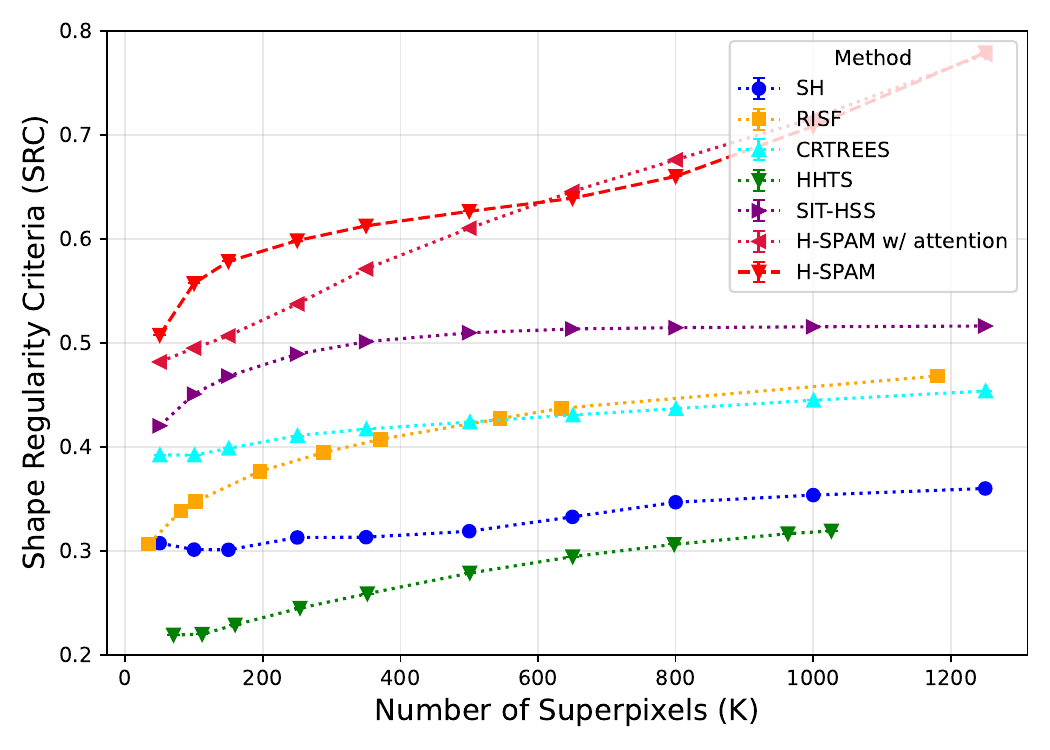}&
\includegraphics[width=0.31\textwidth, height=\heee]{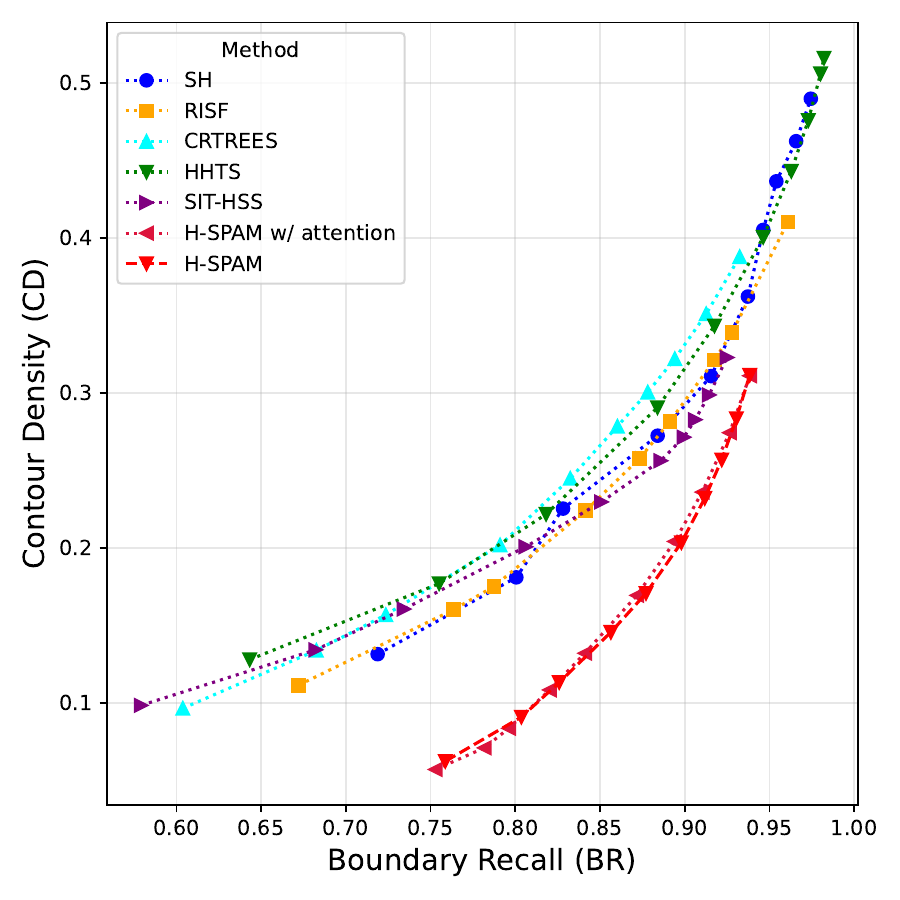}\\

 \rotatebox{90}{\hspace{1cm}{\textbf{NYUv2}}}&

\includegraphics[width=0.31\textwidth, height=\heee]{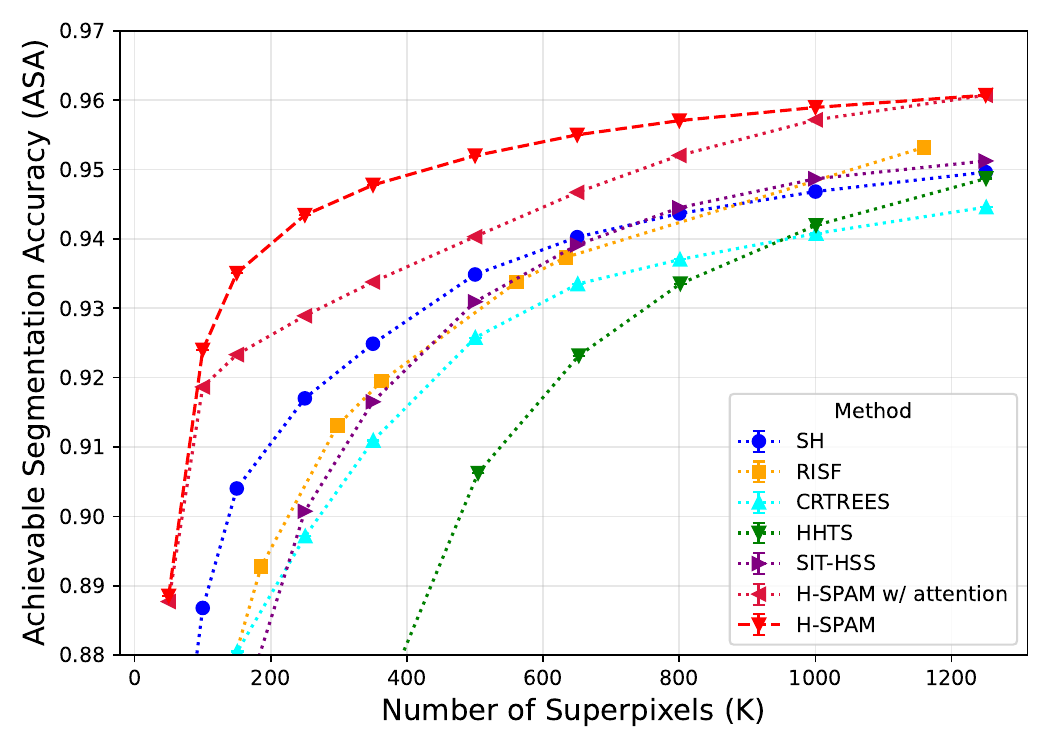}&
\includegraphics[width=0.31\textwidth, height=\heee]{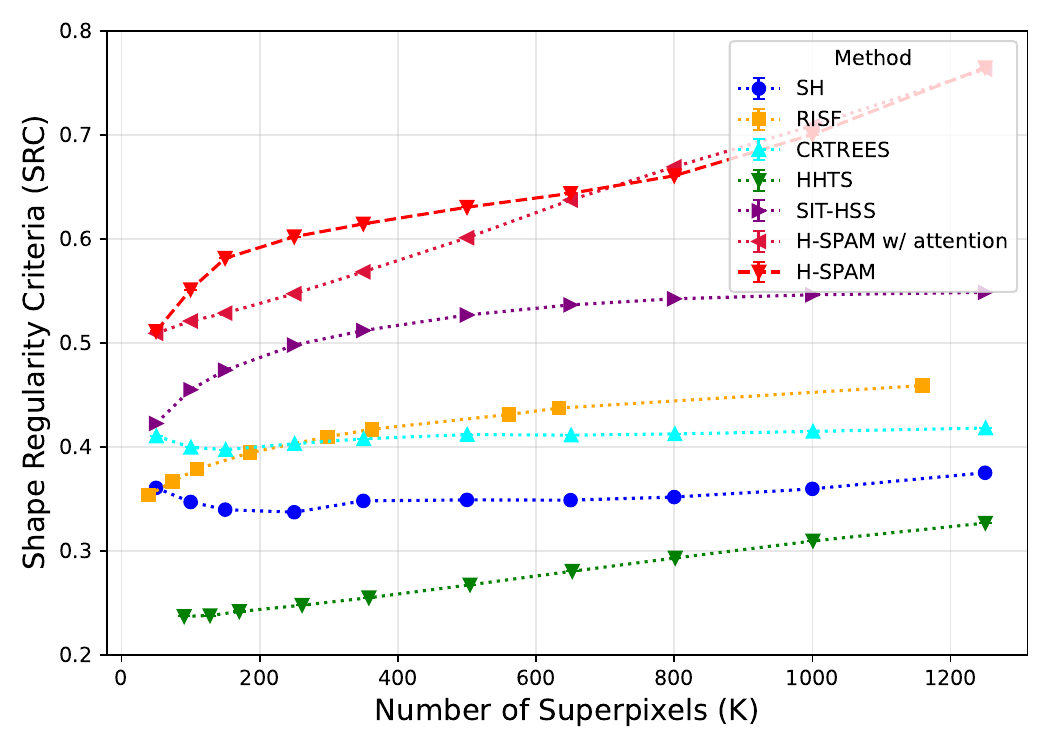}&
\includegraphics[width=0.31\textwidth, height=\heee]{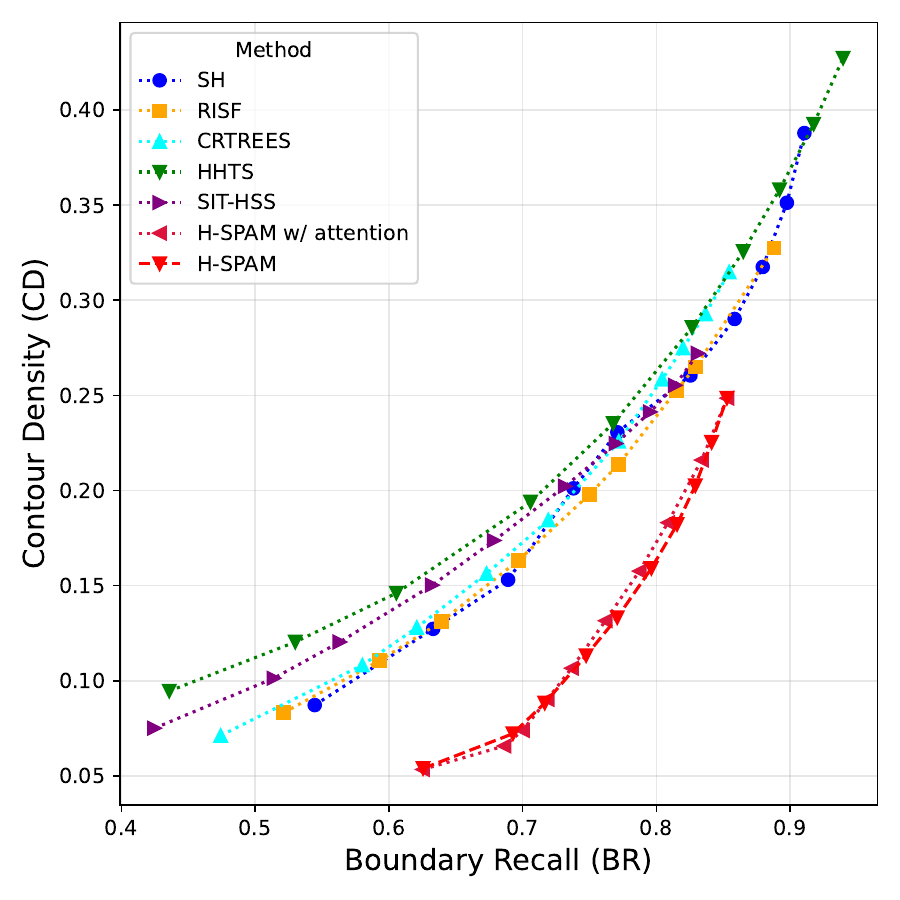}\\

 \rotatebox{90}{\hspace{1cm}{\textbf{SBD}}}&
\includegraphics[width=0.31\textwidth, height=\heee]{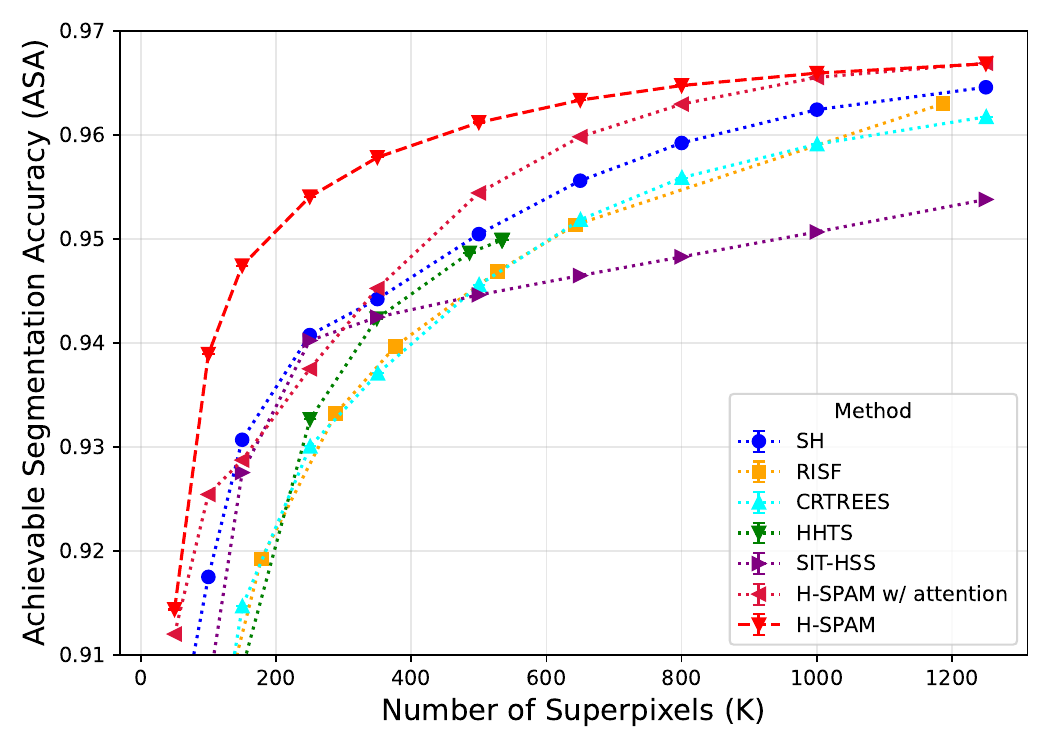}&
\includegraphics[width=0.31\textwidth, height=\heee]{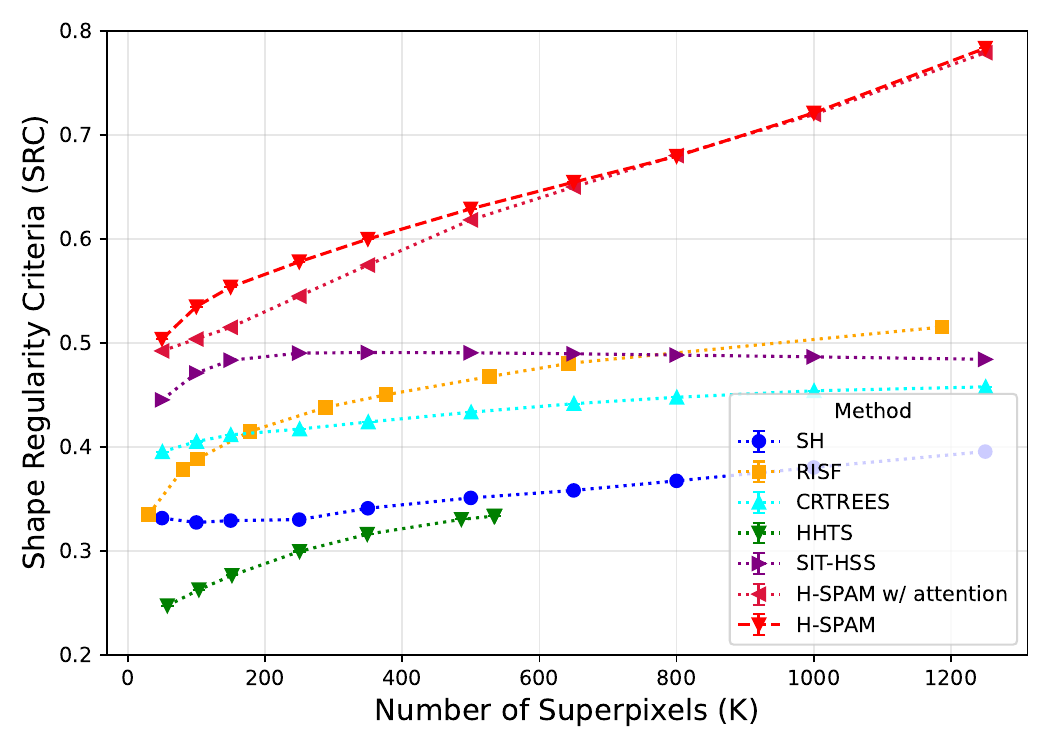}&
\includegraphics[width=0.31\textwidth, height=\heee]{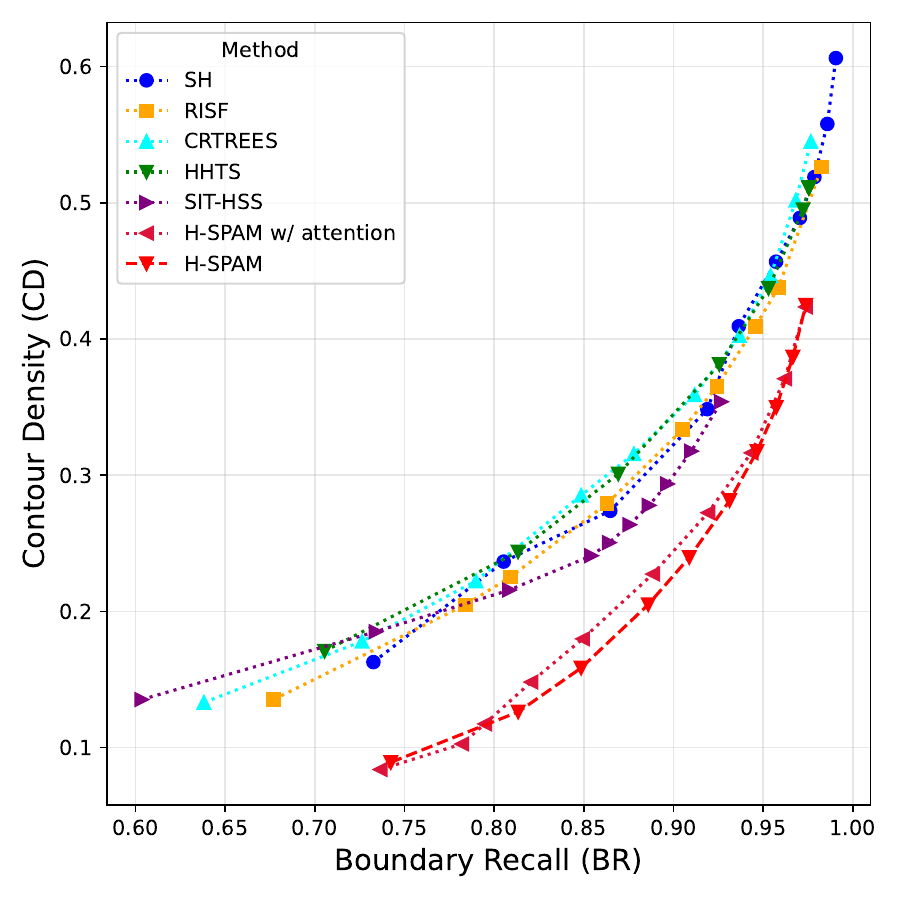}\\[-2ex]
 \end{tabular}}%
\caption{\textbf{Quantitative comparison of H-SPAM to state-of-the-art hierarchical superpixel methods.} 
H-SPAM is the most accurate while being also the most regular hierarchical methods on the three datasets.}
\label{fig:grid_hierarchical_results}
\end{figure*}

\begin{figure*}[t!]
\centering
\newcommand{\heee}{0.21\textwidth}
{\scriptsize
\begin{tabular}{@{\hspace{0mm}}c@{\hspace{2mm}}c@{\hspace{2mm}}c@{\hspace{0mm}}}
        \includegraphics[width=0.32\textwidth, height=\heee]{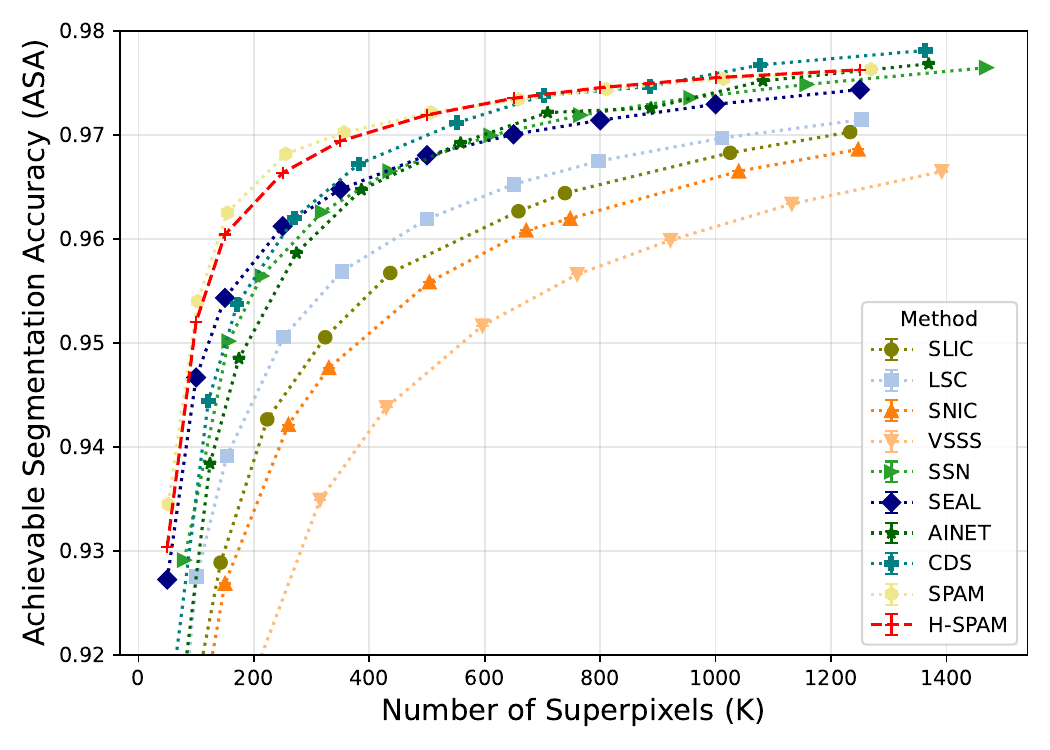}&
        \includegraphics[width=0.32\textwidth, height=\heee]{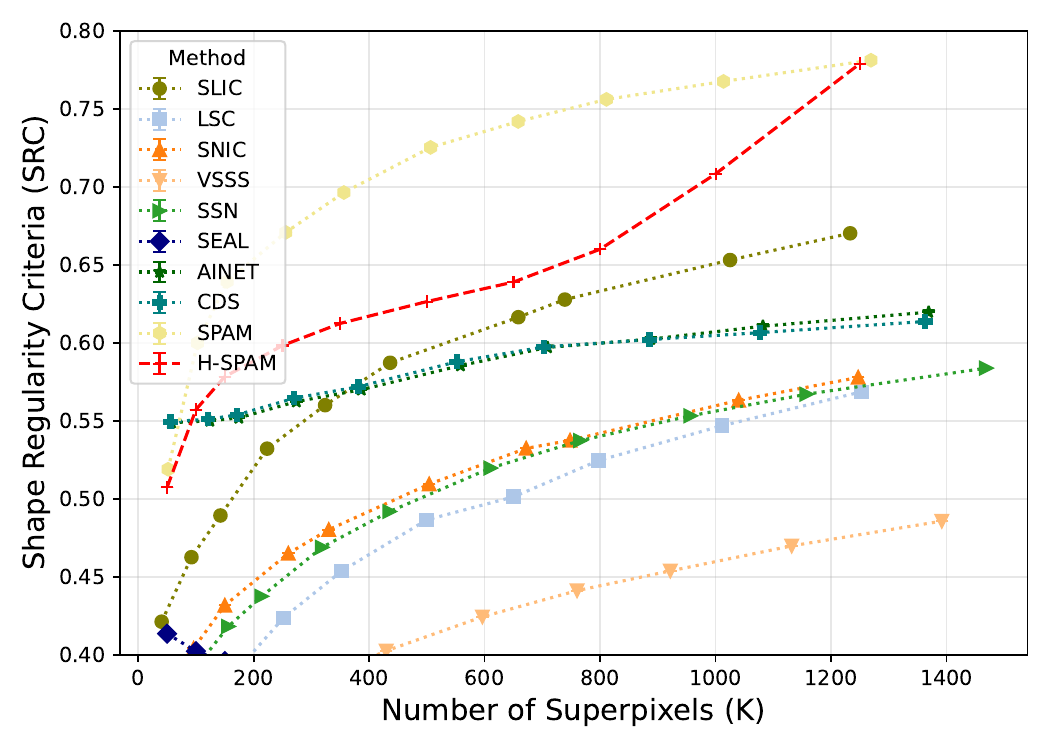}&
        \includegraphics[width=0.32\textwidth, height=\heee]{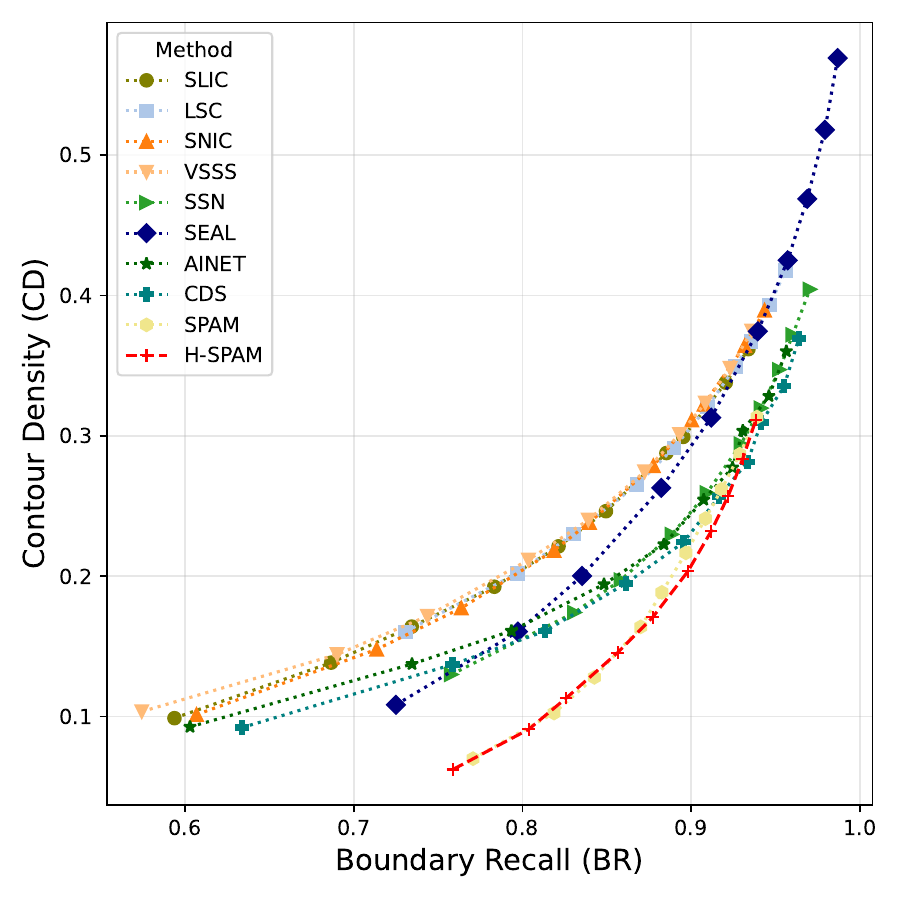}\\[-2ex]
\end{tabular}}
        \label{fig:res_with_all_bsd}
\caption{\textbf{Quantitative comparison of H-SPAM to state-of-the-art non-hierarchical superpixel methods on the BSD.} 
Although constrained to be hierarchical, 
H-SPAM is one of the best performing method in both accuracy and regularity, preserving the performance of single scale object-based approach \cite{Walther2025spam}.}
\label{fig:bsd_with_all}
\end{figure*}

We compared H-SPAM to hierarchical superpixel methods:
SH \cite{wei2018}, 
RISF \cite{galvao2020image},
CRTREES~\cite{yan2022hierarchical},
HHTS \cite{chang2024hierarchical}, and 
SIT-HSS \cite{xie2025hierarchical}.
We also compared against non-hierarchical methods, both traditionnal:
SLIC \cite{achanta2012},
LSC \cite{chen2017},
SNIC \cite{achanta2017superpixels},
VSSS \cite{zhou2023vine}
and deep learning-based:
SSN \cite{jampani2018superpixel},
SEAL \cite{tu2018learning},
AINET \cite{wang2021ainet},
CDS \cite{xu2024learning}
and 
SPAM \cite{Walther2025spam}.
Figure \ref{fig:grid_hierarchical_results} reports the comparison with state-of-the-art hierarchical approaches on the three datasets introduced in Section \ref{subsec:validation}. Across all metrics and over the full range of superpixel numbers $K$, H-SPAM largely outperforms the other hierarchical methods in terms of accuracy (ASA), regularity (SRC), and contour detection (CD/BR).
These trends are consistent across all three datasets, demonstrating the robustness and stability of the proposed method. 
Moreover H-SPAM is able to generate exactly the requested superpixel number.
Finally, we compared H-SPAM with classical non-hierarchical superpixel methods. As shown in Figure~\ref{fig:bsd_with_all}, H-SPAM achieves performance close to SPAM, even though SPAM does not enforce a hierarchy. In terms of ASA and SRC, H-SPAM ranks second after SPAM. The slight performance drop can be explained by the constraint of building a true hierarchy, since early merge decisions may introduce small errors that propagate to higher levels. For the CD/BR ratio, the performance of H-SPAM is almost identical to that of SPAM. Overall, H-SPAM remains the most accurate and regular hierarchical superpixel method.

\subsection{Qualitative evaluation}

\begin{figure*}[t]
\centering
\begin{tabular}{@{}cccc@{}}
    \includegraphics[width=0.24\linewidth]{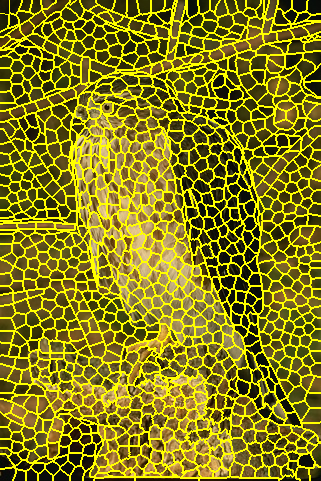}&
    \includegraphics[width=0.24\linewidth]{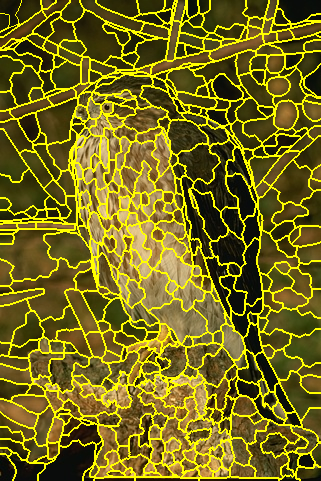}&
    \includegraphics[width=0.24\linewidth]{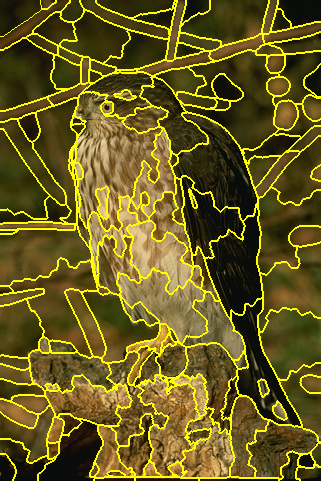}&
    \includegraphics[width=0.24\linewidth]{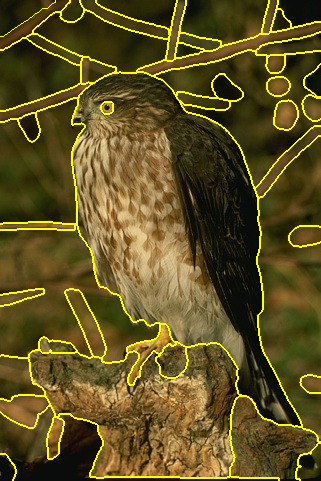}\\[-1.5ex]
\end{tabular}
\caption{\textbf{Qualitative example of H-SPAM} for different scales. From left to right: 1250, 500, 150, 50  superpixels. 
H-SPAM produces perfectly nested, very regular and easily interpretable regions, that align well with object boundaries.
}
\label{fig:qualitative_grid}
\end{figure*}

\begin{figure}[t]
\centering
{\scriptsize
\begin{tabular}{cccc}
\includegraphics[width=0.24\textwidth]{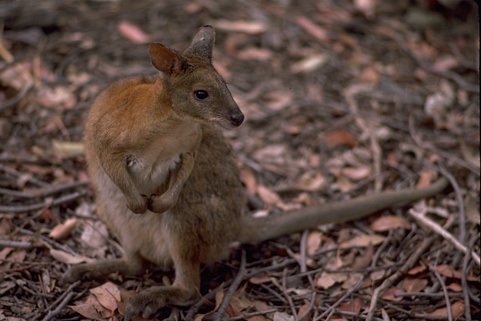} &
\includegraphics[width=0.24\textwidth]{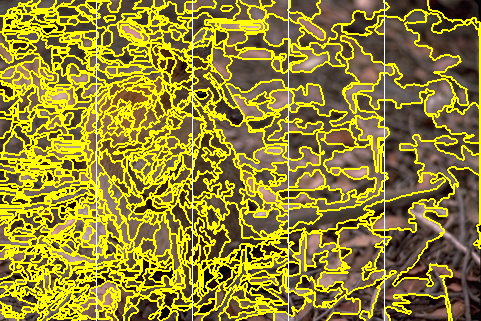} &
\includegraphics[width=0.24\textwidth]{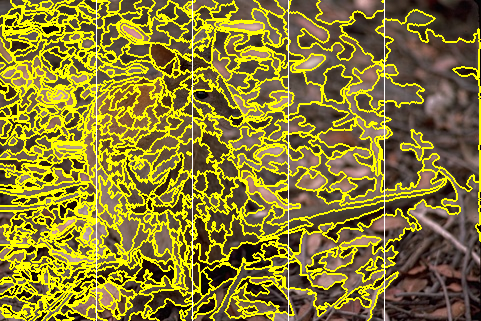} &
\includegraphics[width=0.24\textwidth]{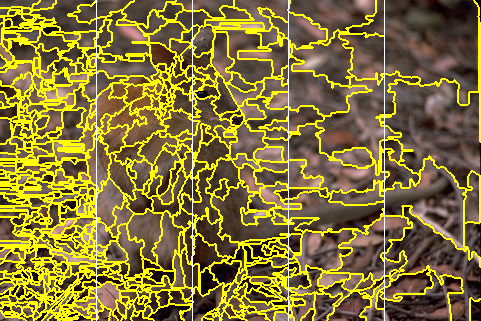} \\
Image &
SH~\cite{wei2018}&
RISF~\cite{galvao2020image} &
CRTREES~\cite{yan2022hierarchical} \\[1ex]
\includegraphics[width=0.24\textwidth]{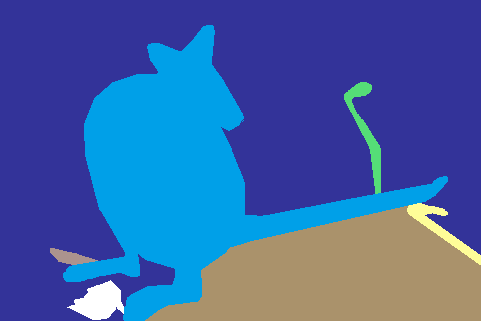} &
\includegraphics[width=0.24\textwidth]{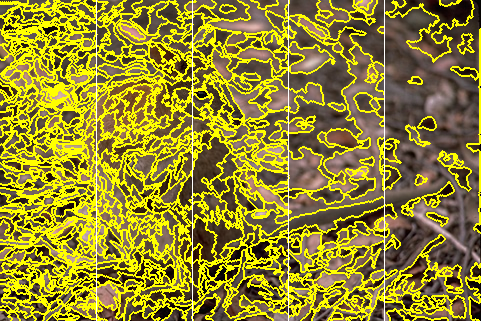} &
\includegraphics[width=0.24\textwidth]{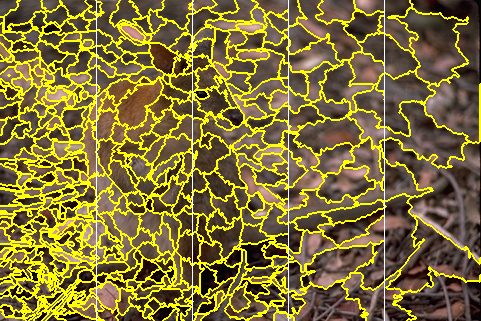} &
\includegraphics[width=0.24\textwidth]{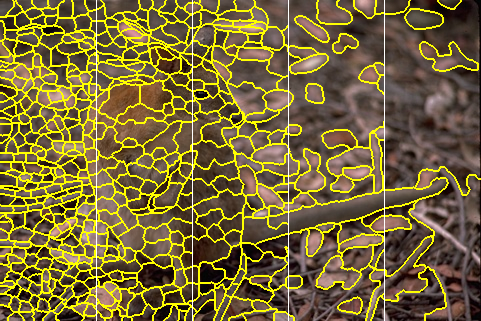} \\
Groundtruth &
HHTS\cite{chang2024hierarchical} &
SIT-HSS~\cite{xie2025hierarchical} &
\textbf{H-SPAM}\\[1ex]
\includegraphics[width=0.24\textwidth]{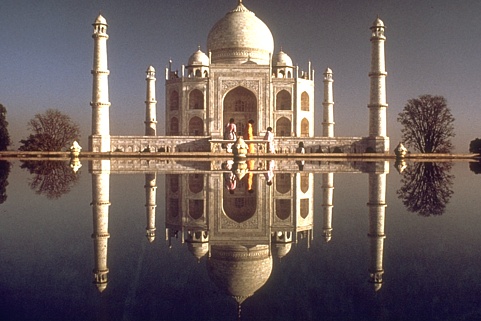} &
\includegraphics[width=0.24\textwidth]{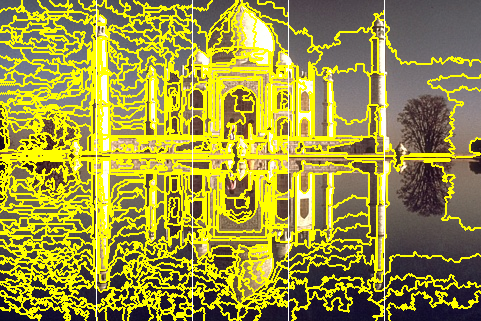} &
\includegraphics[width=0.24\textwidth]{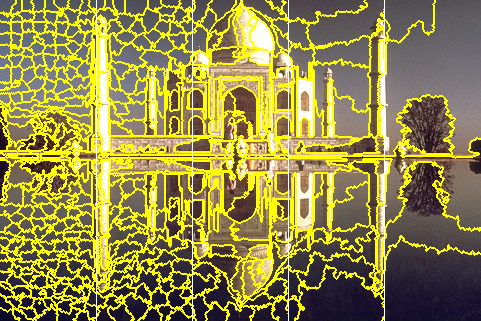} &
\includegraphics[width=0.24\textwidth]{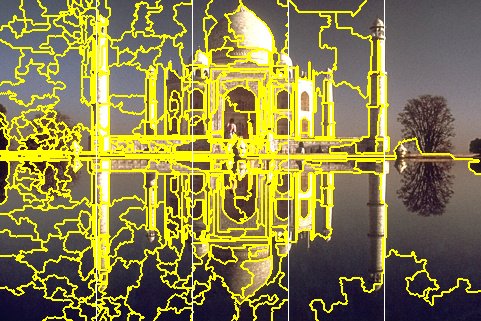} \\
Image &
SH~\cite{wei2018}&
RISF~\cite{galvao2020image} &
CRTREES~\cite{yan2022hierarchical} \\[1ex]
\includegraphics[width=0.24\textwidth]{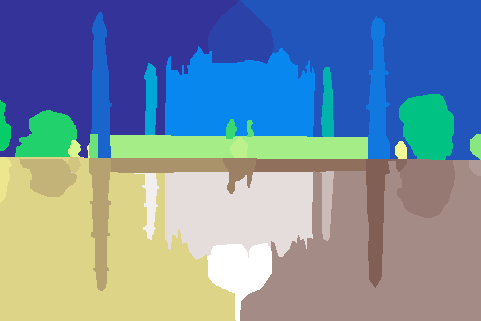} &
\includegraphics[width=0.24\textwidth]{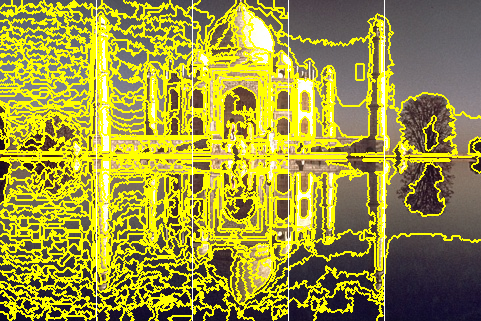} &
\includegraphics[width=0.24\textwidth]{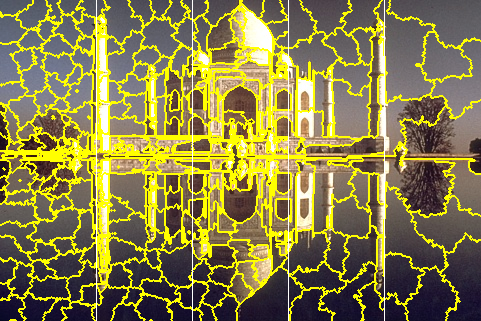} &
\includegraphics[width=0.24\textwidth]{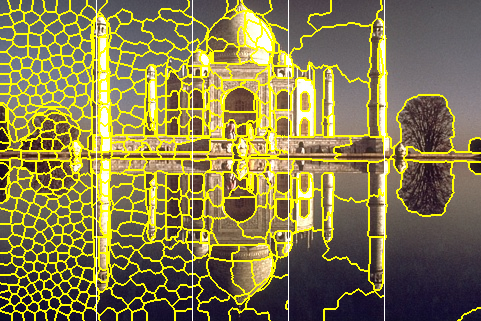} \\
Groundtruth &
HHTS\cite{chang2024hierarchical} &
SIT-HSS~\cite{xie2025hierarchical} &
\textbf{H-SPAM}\\[-1ex]
\end{tabular}
}
\caption{\textbf{Qualitative comparison between hierarchical methods.} Number of superpixels from left to right: 1250, 800, 500, 150, 50.
H-SPAM produces regular and easily interpretable regions, that align well with object boundaries
compared to other methods.
}
\label{fig:sota_grid}
\end{figure}

We report some visual examples in both Figure~\ref{fig:qualitative_grid} and 
Figure~\ref{fig:sota_grid}. 
We observe that all methods 
SH \cite{wei2018}, 
RISF \cite{galvao2020image}
CRTREES~\cite{yan2022hierarchical},
HHTS \cite{chang2024hierarchical}, and 
SIT-HSS \cite{xie2025hierarchical} produce superpixels that are noisy, sometimes difficult to interpret, and less accurate at preserving true object boundaries.
In contrast, H-SPAM builds a true superpixel hierarchy while achieving the highest accuracy and regularity.

\section{Conclusion}

In this work, we introduced the Hierarchical Superpixel Anything Model 
(H-SPAM), a unified method that can produce an accurate, easily interpretable and perfectly nested superpixel hierarchy. 
H-SPAM largely outperforms other hierarchical state-of-the-art methods in both accuracy and regularity.
Our object-based framework enables to preserve the respect of object boundaries 
through a two-phase merging process.
Hence, even with the nestedness constraint, 
H-SPAM remains on par with the best 
 non-hierarchical deep learning-based methods.

With H-SPAM, we further demonstrate the interest of such object-based approach, providing a finer superpixel partition to overcome the limitations of foundation segmentation models.
We also propose alternative segmentation modes using visual attention and user-based interactions to focus the information on key objects.
This contributes to provide a diverse multi-scale segmentation tool 
that would be useful for downstream vision tasks and annotation pipelines.

{
	\small
	\bibliographystyle{splncs04}
	\bibliography{main}
}

\newcolumntype{M}[1]{>{\centering\arraybackslash}m{#1}}
\title{H-SPAM: Hierarchical Superpixel \\Anything Model\\
--- Supplementary Material ---}

\titlerunning{-- Supp. Mat. -- H-SPAM: Hierarchical Superpixel Anything Model}
\author{Julien Walther$^{\;1, \;2}$ \and Rémi Giraud$^{\;1}$ \and Michaël Clément$^{\;2}$ }
\authorrunning{J.Walther et al.}
\institute{Univ. Bordeaux, CNRS, Bordeaux INP, IMS, UMR 5218, France  \and
Univ. Bordeaux, CNRS, Bordeaux INP, LaBRI, UMR 5800, France
}
\maketitle

\section{Impact of initial number of superpixels}
\begin{figure}[t]
\centering
{\scriptsize
\begin{tabular}{cc}
\includegraphics[width=0.49\linewidth]{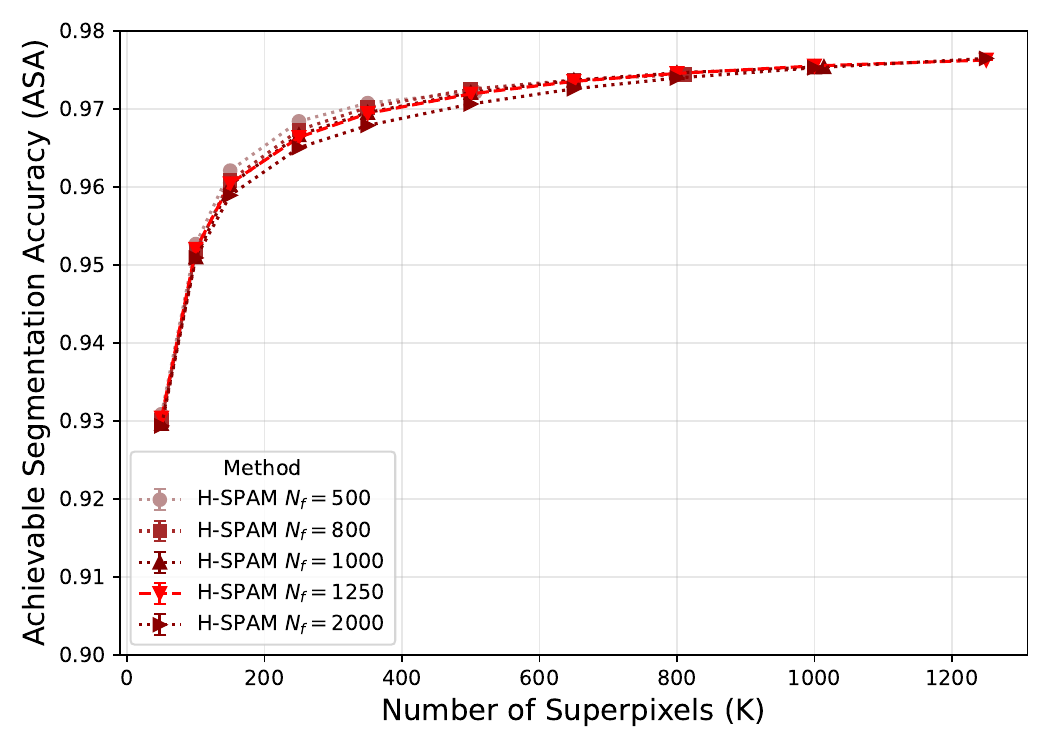} &
\includegraphics[width=0.49\linewidth]{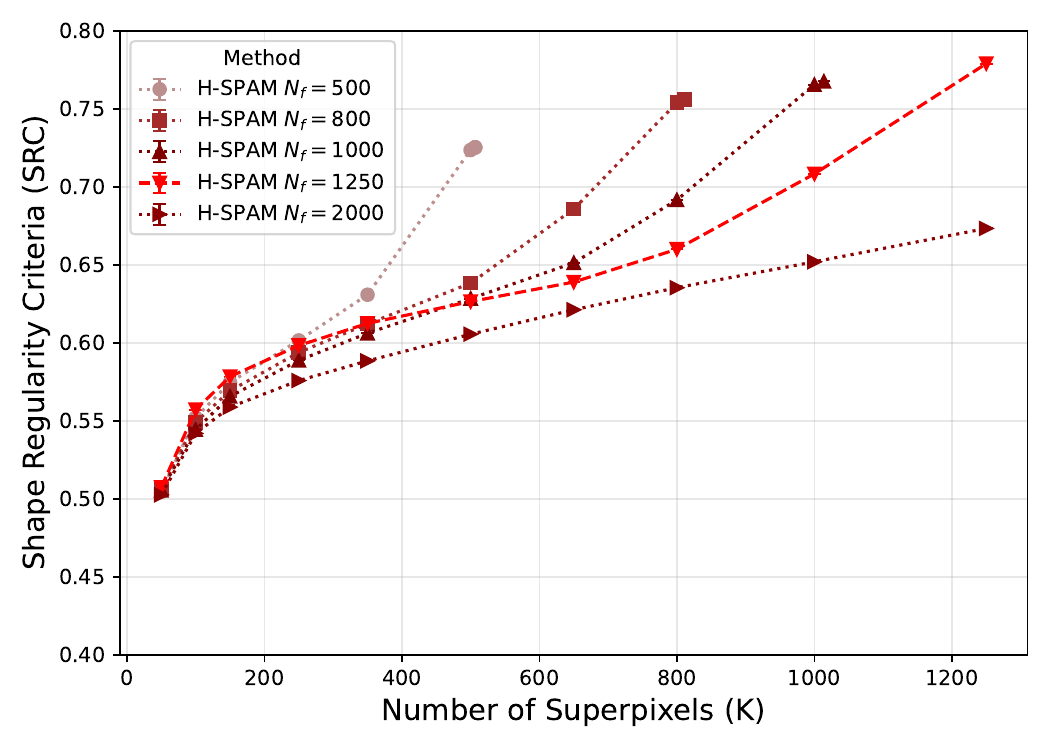} \\
\end{tabular}
}
\caption{\textbf{Influence of the number of superpixel:} Lower $N_f$ performs better at small $K$, while higher $N_f$ becomes better at intermediate $K$.}
\label{fig:ablation_nb_sp_in}
\end{figure}

Figure \ref{fig:ablation_nb_sp_in} illustrates the influence of the number of superpixels in the initial map on the constructed hierarchy. We observe that an initialization with fewer superpixels performs better at low values of K, while initializations with a larger number of superpixels become more favorable at intermediate values, leading to crossings between the curves. This behavior can be explained by the role of the initialization in the depth of the hierarchy. As the initial number of superpixels increases, the hierarchy must go through more merge steps to reach a given K, which allows small early fusion errors to propagate and reduces accuracy at that level. In practice, the choice of initialization has a direct impact on performance and should be set to the higher number wanted. In our experiments, tests were conducted using an initialization of 1250 superpixels.

\begin{figure}[t]
\newcommand{\hhh}{0.225\textwidth}
\centering
{\scriptsize
\begin{tabular}{cccccc}
&\rotatebox{90}{\textbf{\text{ }w/o objects}}
&\includegraphics[width=\hhh]{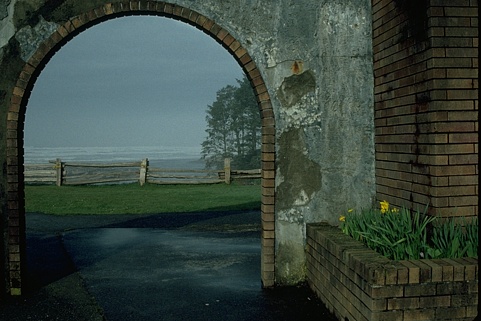} &
\includegraphics[width=\hhh]{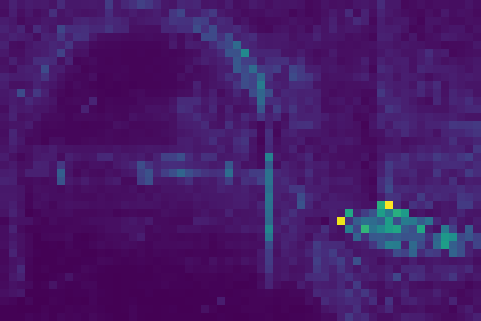} &
\includegraphics[width=\hhh]{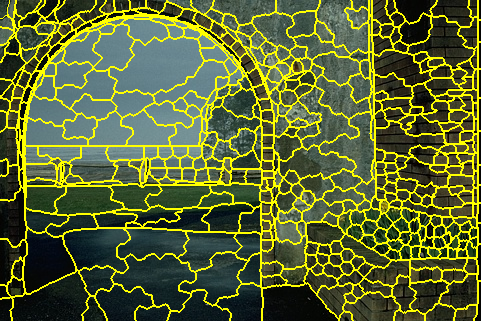} &
\includegraphics[width=\hhh]{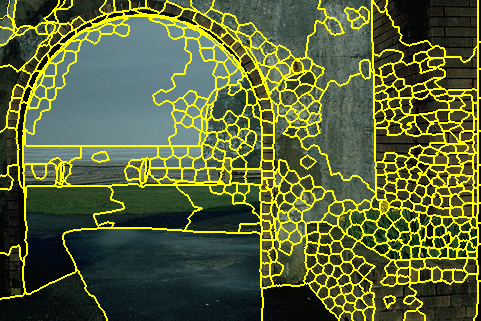} \\
&\rotatebox{90}{\textbf{\text{ }w/ objects}}
&\includegraphics[width=\hhh]{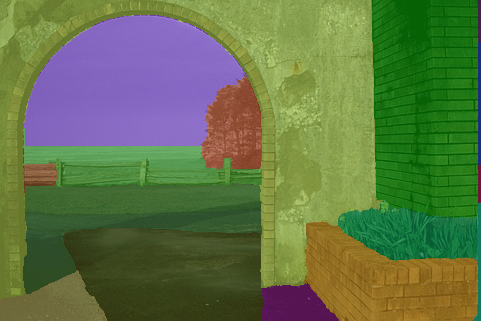} &
\includegraphics[width=\hhh]{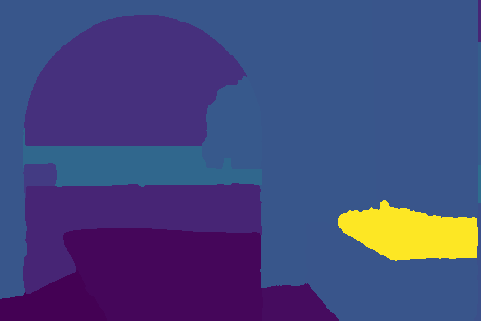} &
\includegraphics[width=\hhh]{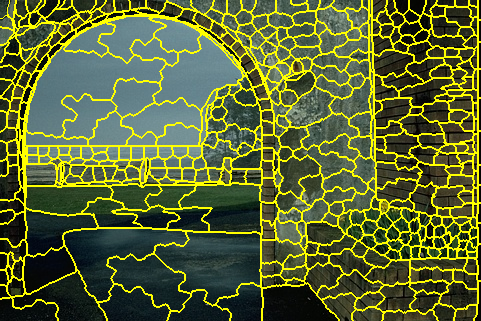} &
\includegraphics[width=\hhh]{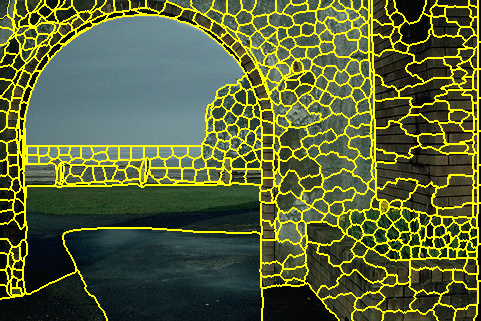} \\
\rotatebox{90}{\textbf{\text{ }w/ objects}}&\rotatebox{90}{\textbf{\text{ }\text{ }\text{ }+clicks}}
&\includegraphics[width=\hhh]{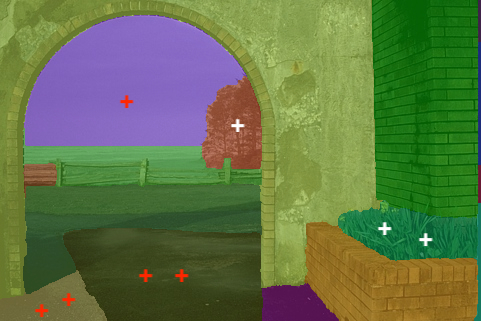} &
\includegraphics[width=\hhh]{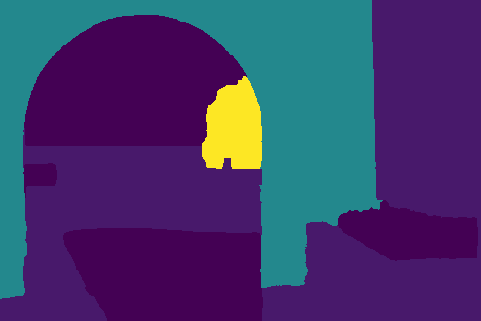} &
\includegraphics[width=\hhh]{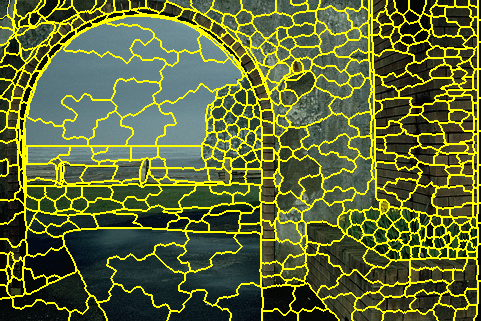} &
\includegraphics[width=\hhh]{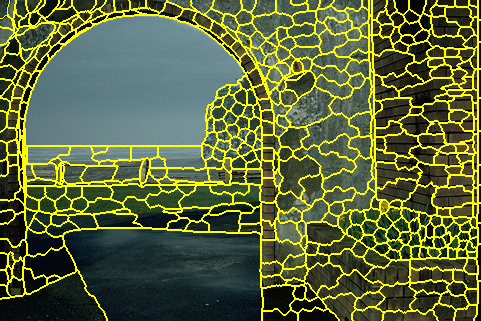} \\
&&Object overlay &
Attention map &
$w_{att}=0.01$ &
$w_{att}=0.5$
\end{tabular}
}
\caption{\textbf{Illustration of the attention modes}. With our object-based framework (middle row), the attention can be averaged within objects to provide a cleaner guide for the merging process. The bottom row shows the user interactive mode, where red/white crosses lead to fewer/more superpixels in the object.}
\label{fig:Va_attention}
\end{figure}

\section{Additional Qualitative Examples}
\begin{figure}[t]
\centering
{\scriptsize
\begin{tabular}{cccc}
\includegraphics[width=0.24\textwidth]{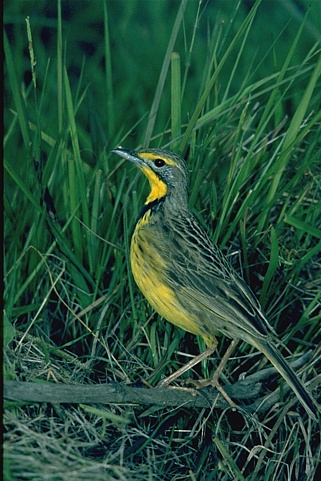} &
\includegraphics[width=0.24\textwidth]{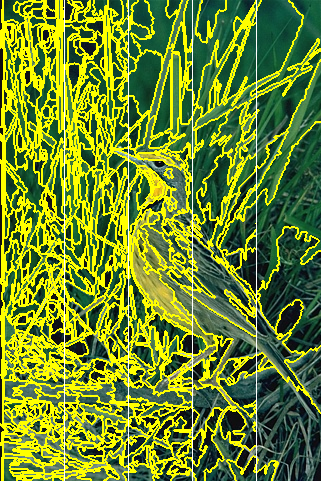} &
\includegraphics[width=0.24\textwidth]{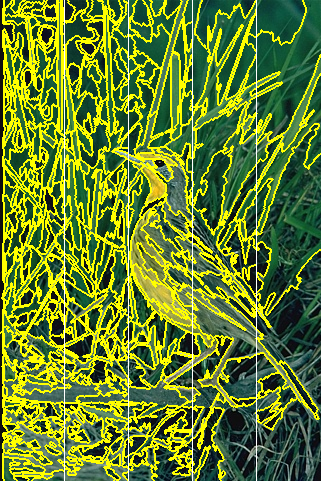} &
\includegraphics[width=0.24\textwidth]{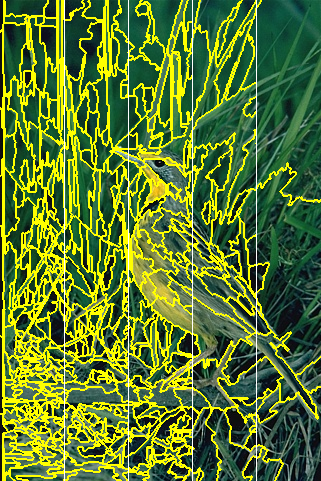} \\
Image &
SH~\cite{wei2018}&
RISF~\cite{galvao2020image} &
CRTREES~\cite{yan2022hierarchical} \\[1ex]
\includegraphics[width=0.24\textwidth]{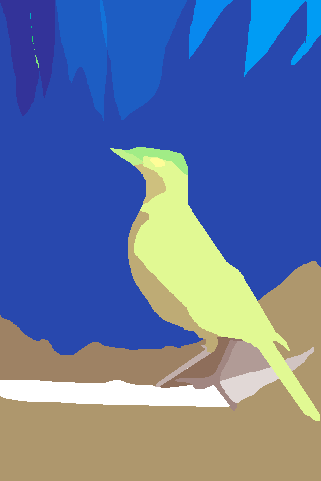} &
\includegraphics[width=0.24\textwidth]{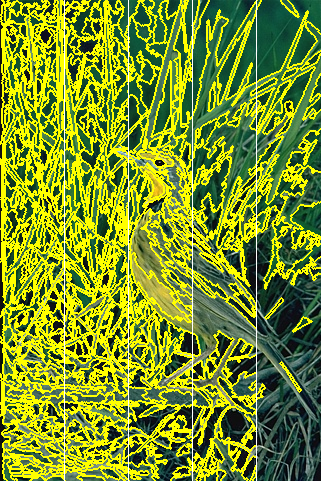} &
\includegraphics[width=0.24\textwidth]{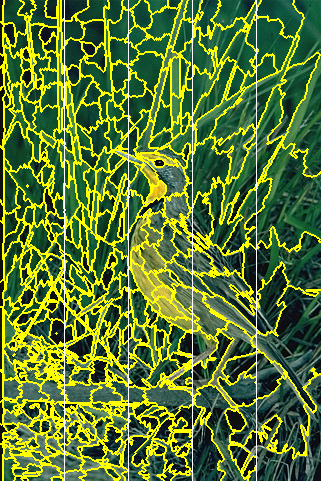} &
\includegraphics[width=0.24\textwidth]{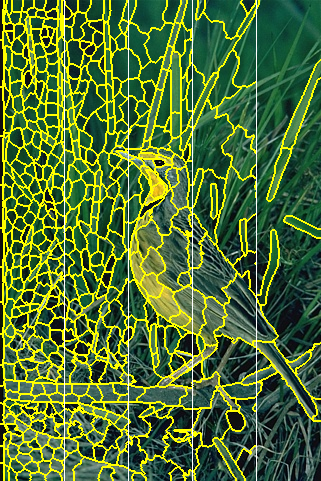} \\
GroundTruth &
HHTS\cite{chang2024hierarchical} &
SIT-HSS~\cite{xie2025hierarchical} &
\textbf{H-SPAM}\\[1ex]

\includegraphics[width=0.24\textwidth]{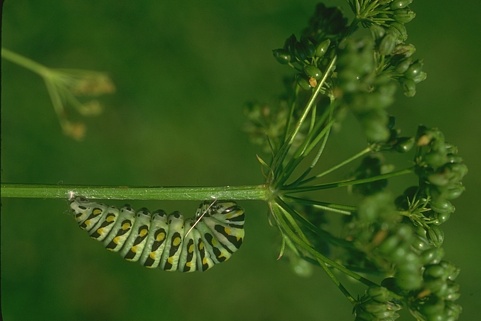} &
\includegraphics[width=0.24\textwidth]{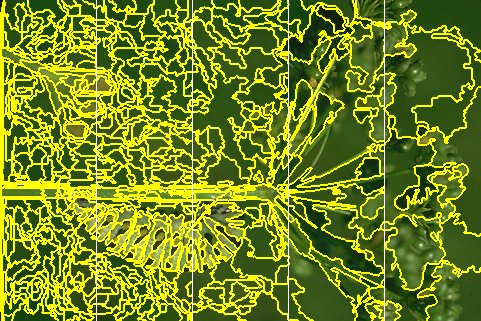} &
\includegraphics[width=0.24\textwidth]{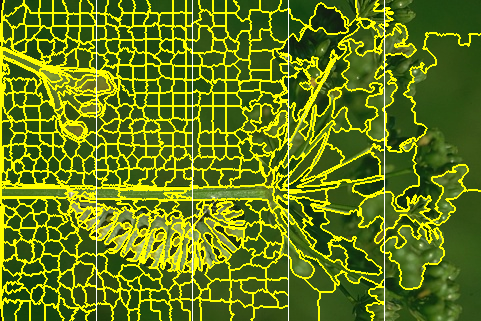} &
\includegraphics[width=0.24\textwidth]{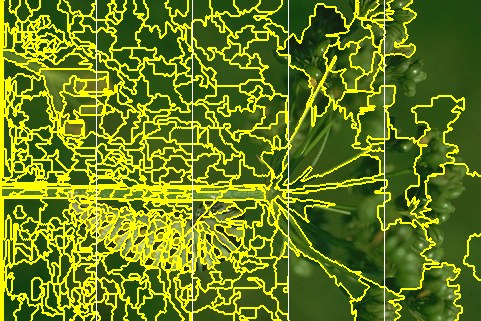} \\
Image &
SH~\cite{wei2018}&
RISF~\cite{galvao2020image} &
CRTREES~\cite{yan2022hierarchical} \\[1ex]
\includegraphics[width=0.24\textwidth]{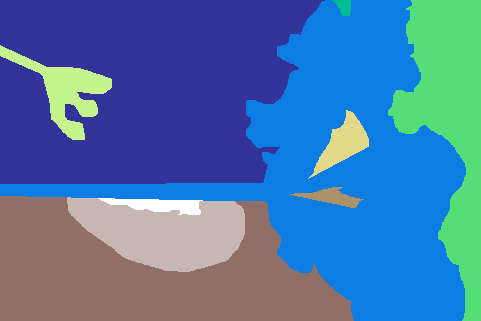} &
\includegraphics[width=0.24\textwidth]{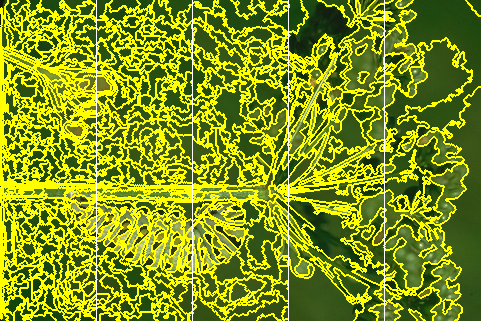} &
\includegraphics[width=0.24\textwidth]{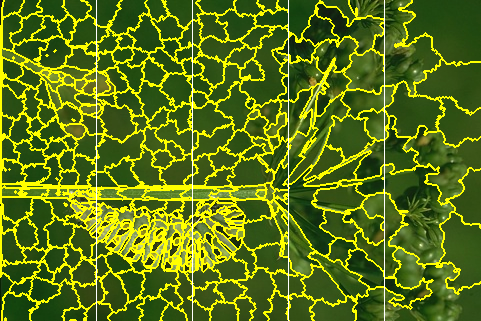} &
\includegraphics[width=0.24\textwidth]{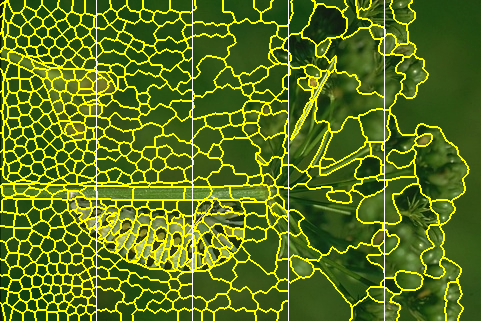} \\
GroundTruth &
HHTS\cite{chang2024hierarchical} &
SIT-HSS~\cite{xie2025hierarchical} &
\textbf{H-SPAM}\\[1ex]

\end{tabular}
}
\caption{\textbf{Qualitative comparison between hierarchical methods.} Number of superpixels from left to right: 1250, 800, 500, 150, 50.
H-SPAM produces regular and easily interpretable regions, that align well with object boundaries compared to other methods.
}
\label{fig:comparaison_1}
\end{figure}

\begin{figure}[t]
\centering
{\scriptsize
\begin{tabular}{cccc}

\includegraphics[width=0.24\textwidth]{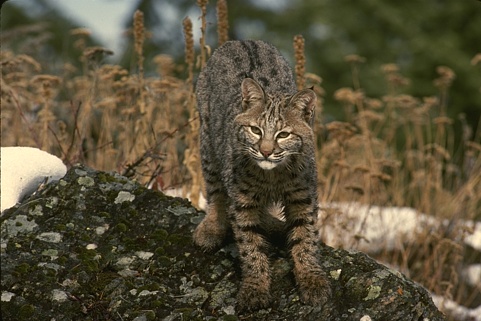} &
\includegraphics[width=0.24\textwidth]{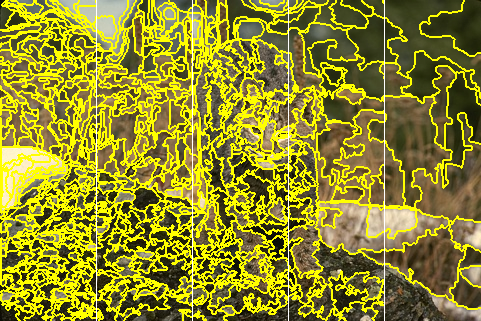} &
\includegraphics[width=0.24\textwidth]{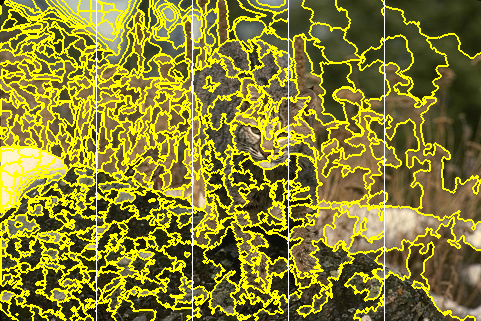} &
\includegraphics[width=0.24\textwidth]{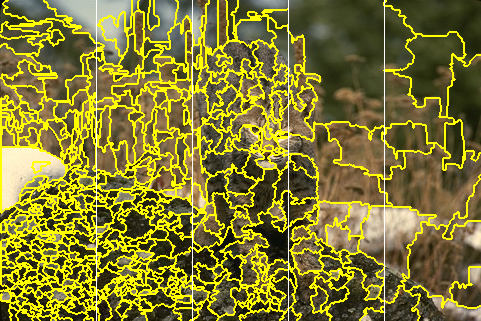} \\
Image &
SH~\cite{wei2018}&
RISF~\cite{galvao2020image} &
CRTREES~\cite{yan2022hierarchical} \\[1ex]
\includegraphics[width=0.24\textwidth]{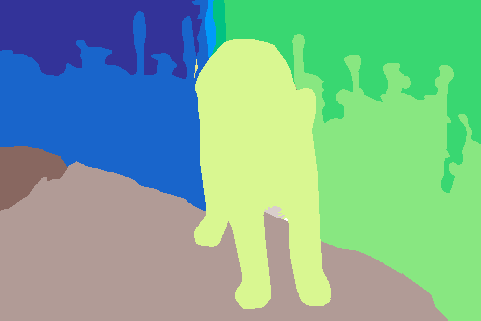} &
\includegraphics[width=0.24\textwidth]{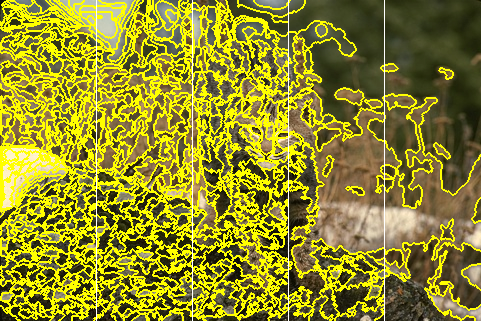} &
\includegraphics[width=0.24\textwidth]{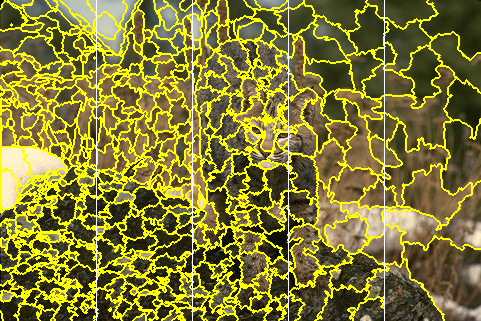} &
\includegraphics[width=0.24\textwidth]{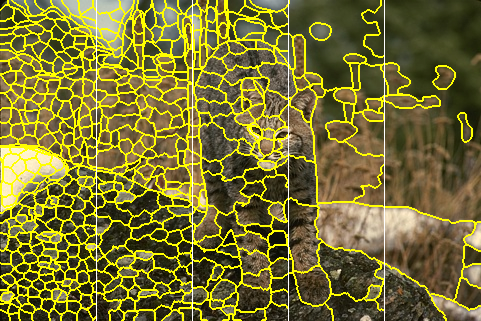} \\
GroundTruth &
HHTS\cite{chang2024hierarchical} &
SIT-HSS~\cite{xie2025hierarchical} &
\textbf{H-SPAM}\\[1ex]

\includegraphics[width=0.24\textwidth]{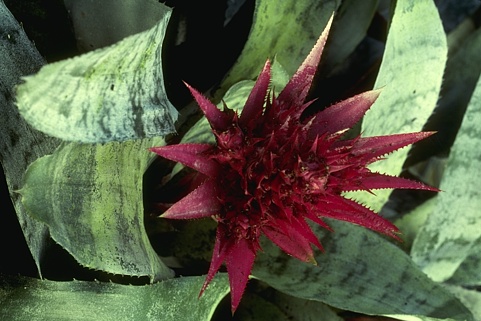} &
\includegraphics[width=0.24\textwidth]{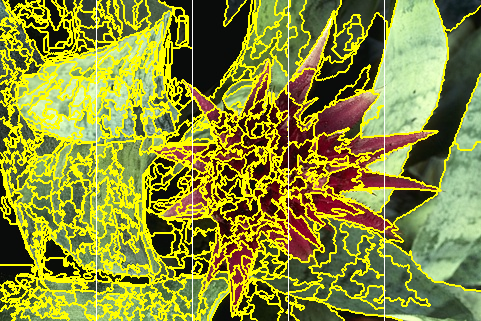} &
\includegraphics[width=0.24\textwidth]{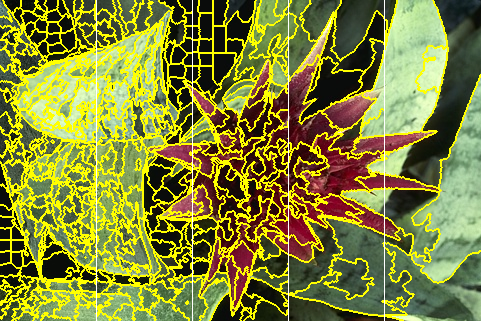} &
\includegraphics[width=0.24\textwidth]{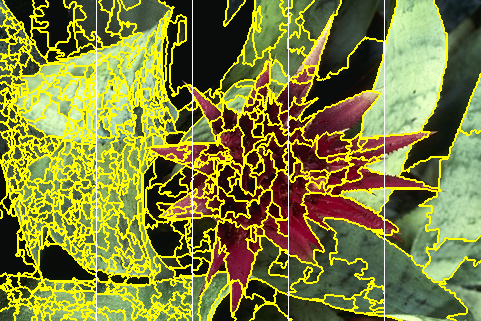} \\
Image &
SH~\cite{wei2018}&
RISF~\cite{galvao2020image} &
CRTREES~\cite{yan2022hierarchical} \\[1ex]
\includegraphics[width=0.24\textwidth]{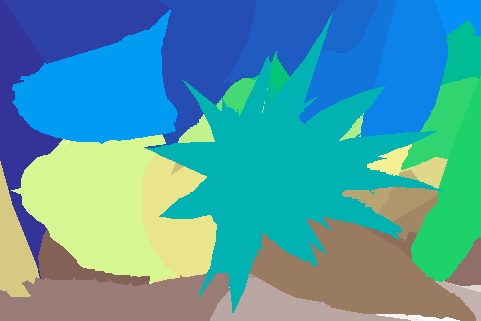} &
\includegraphics[width=0.24\textwidth]{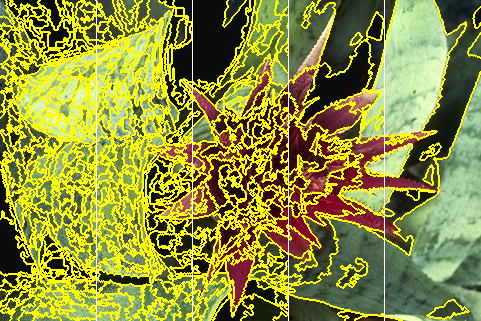} &
\includegraphics[width=0.24\textwidth]{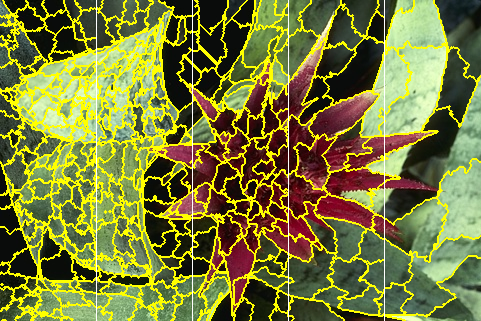} &
\includegraphics[width=0.24\textwidth]{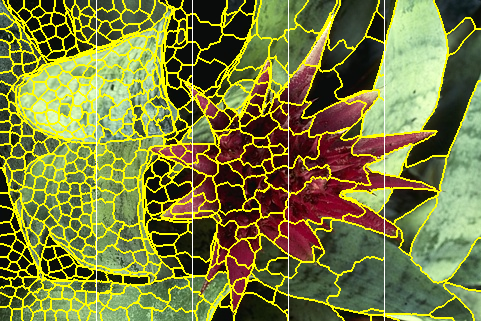} \\
GroundTruth &
HHTS\cite{chang2024hierarchical} &
SIT-HSS~\cite{xie2025hierarchical} &
\textbf{H-SPAM}\\[1ex]

\includegraphics[width=0.24\textwidth]{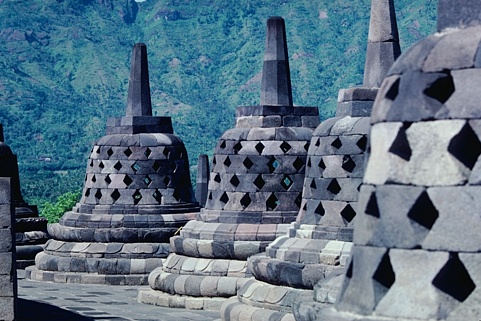} &
\includegraphics[width=0.24\textwidth]{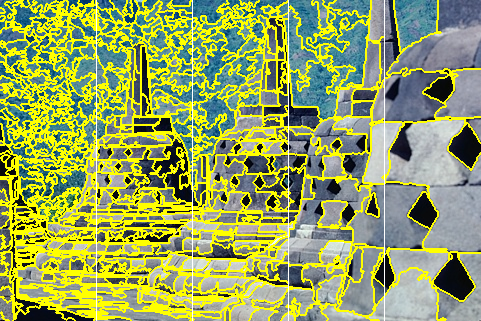} &
\includegraphics[width=0.24\textwidth]{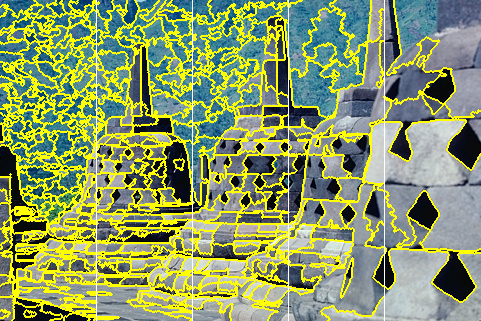} &
\includegraphics[width=0.24\textwidth]{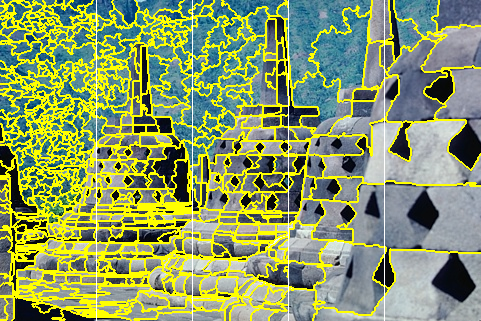} \\
Image &
SH~\cite{wei2018}&
RISF~\cite{galvao2020image} &
CRTREES~\cite{yan2022hierarchical} \\[1ex]
\includegraphics[width=0.24\textwidth]{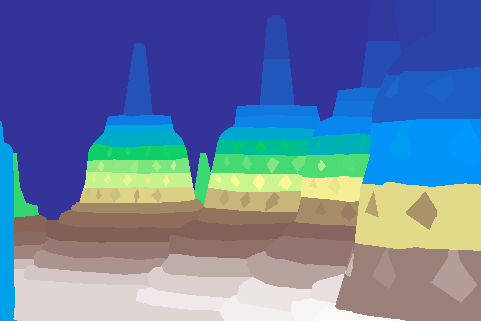} &
\includegraphics[width=0.24\textwidth]{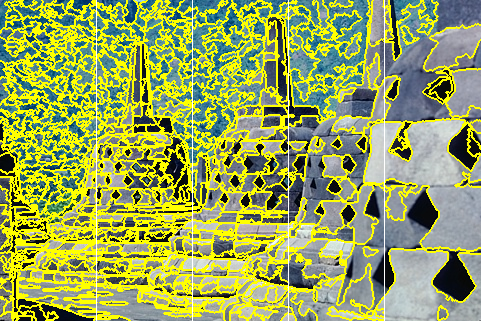} &
\includegraphics[width=0.24\textwidth]{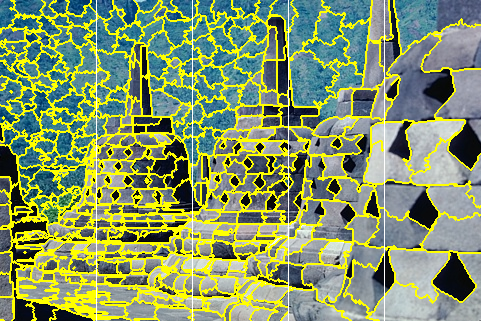} &
\includegraphics[width=0.24\textwidth]{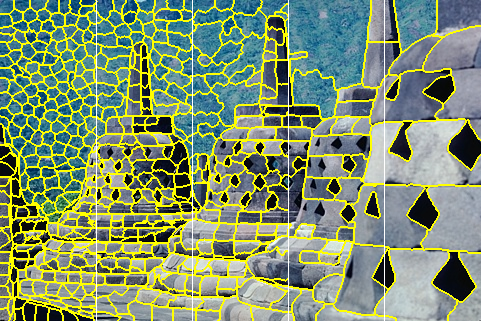} \\
GroundTruth &
HHTS\cite{chang2024hierarchical} &
SIT-HSS~\cite{xie2025hierarchical} &
\textbf{H-SPAM}\\[-1ex]
\end{tabular}
}
\caption{\textbf{Qualitative comparison between hierarchical methods.} Number of superpixels from left to right: 1250, 800, 500, 150, 50.
H-SPAM produces regular and easily interpretable regions, that align well with object boundaries
compared to other methods..
}
\label{fig:comparaison_2}
\end{figure}

\begin{figure*}[t]
\centering
\begin{tabular}{@{}cccc@{}}
    \includegraphics[width=0.24\linewidth]{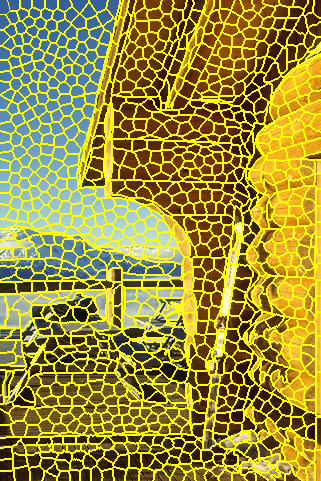}&
    \includegraphics[width=0.24\linewidth]{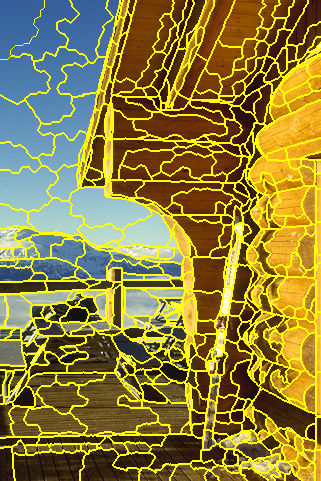}&
    \includegraphics[width=0.24\linewidth]{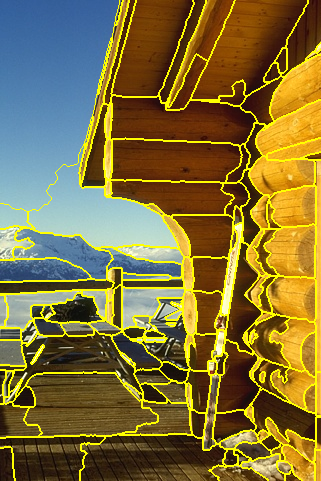}&
    \includegraphics[width=0.24\linewidth]{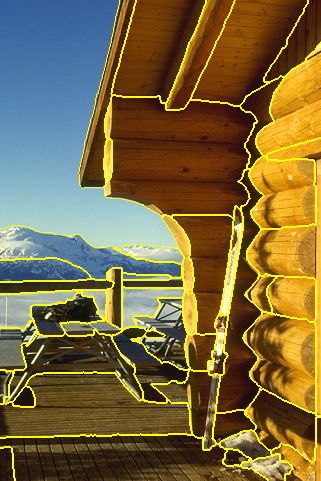}\\

    \includegraphics[width=0.24\linewidth]{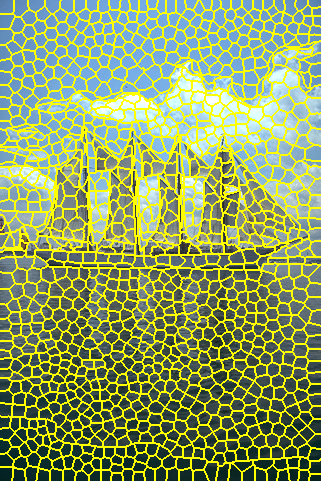}&
    \includegraphics[width=0.24\linewidth]{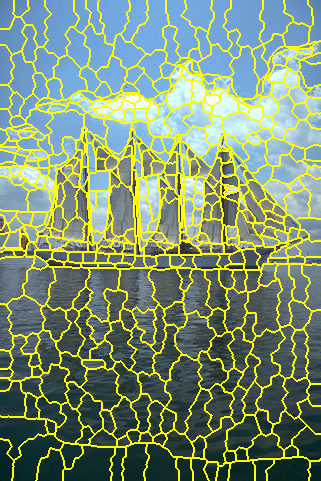}&
    \includegraphics[width=0.24\linewidth]{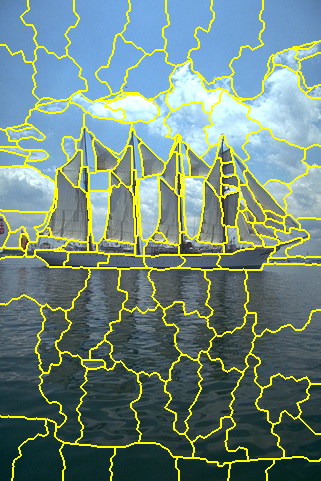}&
    \includegraphics[width=0.24\linewidth]{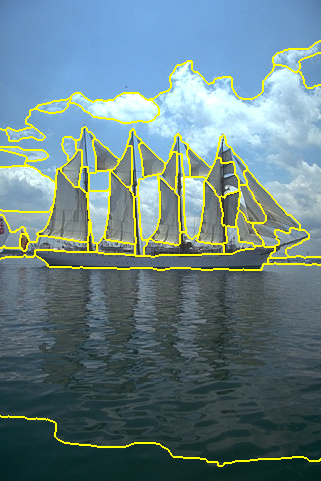}\\
    \includegraphics[width=0.24\linewidth]{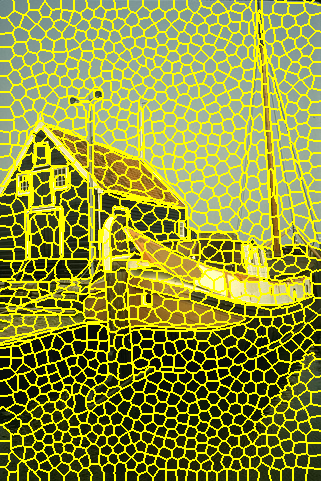}&
    \includegraphics[width=0.24\linewidth]{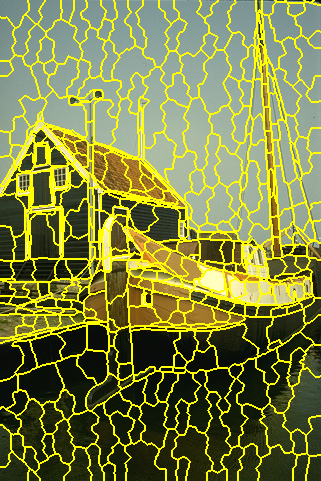}&
    \includegraphics[width=0.24\linewidth]{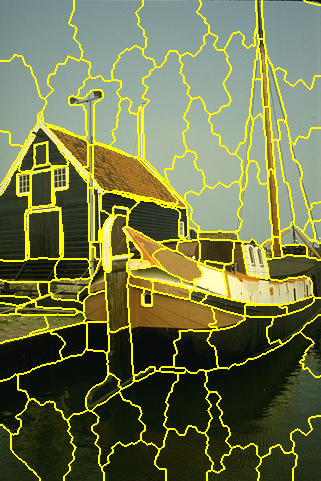}&
    \includegraphics[width=0.24\linewidth]{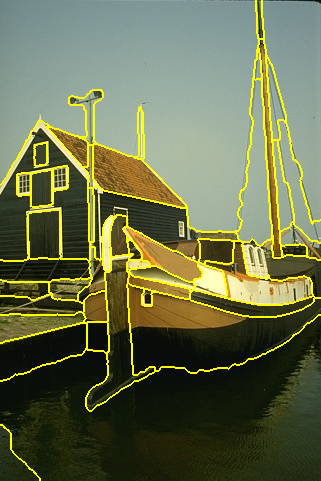}\\

\end{tabular}
\caption{\textbf{Qualitative example of H-SPAM} for different scales. From left to right: 1250, 500, 150, 50  superpixels. 
H-SPAM produces perfectly nested, very regular and easily interpretable regions, that align well with object boundaries compared to other methods.
}
\label{fig:hspam_result_1}
\end{figure*}

In this section, we present additional qualitative results:
examples using the attention mode (Figure~\ref{fig:Va_attention}), 
comparisons with state-of-the-art methods (Figure~\ref{fig:comparaison_1} and Figure~\ref{fig:comparaison_2}), and complete results of H-SPAM at multiple scales (Figure~\ref{fig:hspam_result_1}).

\end{document}